\documentclass[runningheads]{llncs}
\usepackage{graphicx}
\usepackage{comment}
\usepackage{amsmath,amssymb} % define this before the line numbering.
\usepackage{color}

\usepackage[width=122mm,left=12mm,paperwidth=146mm,height=193mm,top=12mm,paperheight=217mm]{geometry}
\usepackage{xspace}
\usepackage{booktabs} % To thicken table lines
\usepackage{bm}

\newcommand{\figLabel}{Fig.\xspace}
\newcommand{\eqLabel}{Equation\xspace}
\newcommand{\secLabel}{Section\xspace}
\newcommand{\tblLabel}{Table\xspace}
\newcommand{\mysection}[1]{\vspace{3pt}\noindent\textbf{#1.}}

\newcommand{\ie}{\textit{i}.\textit{e}.}

\usepackage{multirow}

\newcommand{\B}{\bfseries}
\newcommand{\myNAS}{LC-NAS\xspace}
\begin{document}
% \linenumbers
\pagestyle{headings}
\mainmatter
\def\ECCVSubNumber{2774}  % Insert your submission number here

\title{LC-NAS: Latency Constrained Neural Architecture Search for Point Cloud Networks} % Replace with your title

% %\begin{comment}
% \titlerunning{ECCV-20 submission ID \ECCVSubNumber} 
% \authorrunning{ECCV-20 submission ID \ECCVSubNumber} 
% \author{Anonymous ECCV submission}
% \institute{Paper ID \ECCVSubNumber}
% %\end{comment}
%******************

% \begin{comment}
\titlerunning{LC-NAS}
\author{Guohao Li \and
Mengmeng Xu \and Silvio Giancola \and Ali Thabet \and Bernard Ghanem}
\authorrunning{Li, G. et al.}
\institute{Visual Computing Center,~ KAUST, ~ Thuwal,~ Saudi Arabia 
\\
\email{{\tt\footnotesize \{guohao.li, mengmeng.xu, silvio.giancola, ali.thabet, bernard.ghanem\}@kaust.edu.sa}}}
% \end{comment}
%******************

% \author{Guohao Li\thanks{equal contribution} \quad Chenxin Xiong \footnotemark[1] \quad Ali Thabet\quad Bernard Ghanem\\
% 		Visual Computing Center,~ KAUST \\ ~ Thuwal,~ Saudi Arabia \\
% 		{\tt\footnotesize \{guohao.li, chenxin.xiong, ali.thabet, bernard.ghanem\}@kaust.edu.sa}
% 		}
\maketitle

\begin{abstract}
Point cloud architecture design has become a crucial problem for 3D deep learning. Several efforts exist to manually design architectures with high accuracy in point cloud tasks such as classification, segmentation, and detection. Recent progress in automatic Neural Architecture Search (NAS) minimizes the human effort in network design and optimizes high performing architectures. However, these efforts fail to consider important factors such as latency during inference. Latency is of high importance in time critical applications like self-driving cars, robot navigation, and mobile applications, that are generally bound by the available hardware. In this paper, we introduce a new NAS framework, dubbed \myNAS, where we search for point cloud architectures that are constrained to a target latency. We implement a novel latency constraint formulation to trade-off between accuracy and latency in our architecture search. Contrary to previous works, our latency loss guarantees that the final network achieves latency under a specified target value. This is crucial when the end task is to be deployed in a limited hardware setting. 
Extensive experiments show that \myNAS is able to find state-of-the-art architectures for point cloud classification in ModelNet40 with minimal computational cost. We also show how our searched architectures achieve any desired latency with a reasonably low drop in accuracy. Finally, we show how our searched architectures easily transfer to a different task, part segmentation on PartNet, where we achieve state-of-the-art results while lowering latency by a factor of 10.

\end{abstract}
\section{Introduction}
\begin{figure}[t]
    \centering
    \includegraphics[width=1\textwidth,trim={1cm 4.5cm 1cm 5cm},clip] {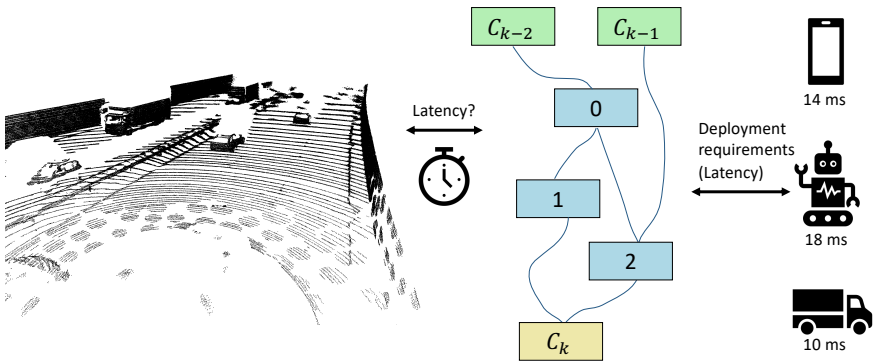}
    \caption{\textbf{\myNAS}. Given a point cloud input task like detecting cars on a LiDAR input (left), we aim to search for the best point cloud architecture, bounded by a latency constraint (middle). This constraint will usually come from deployment requirements. Such requirements will be bound to the deployment hardware, be it a car, navigation robot, or a mobile device (right).}
    \label{fig:intro}
\end{figure}

% Explain why latency nor flops?
Deep learning is the \textit{de facto} choice for solving vision related tasks. In particular, Convolutional Neural Networks (CNNs) achieve state-of-the-art results in 2D image classification \cite{he2016deep,szegedy2016rethinking}, segmentation \cite{chen2017deeplab,ronneberger2015u}, and detection \cite{he2017mask,ren2015faster}. Recent efforts translated this success to 3D point clouds. The seminal PointNet work \cite{pc_qi2017pointnet} opened the way to increasingly performing architectures for processing point clouds. Similar to the 2D case, state-of-the-art architectures achieve impressive gains in point cloud classification \cite{pc_pointcnn}, segmentation \cite{pc_kpconv}, and detection \cite{ding2019votenet,chi2019prcnn}. However, this success comes at the expense of increased computation. Of particular interest is the computational latency during inference, a factor that is paramount to time-critical applications like self-driving cars, robot navigation, and mobile applications (\figLabel \ref{fig:intro}). These applications require fast decision making and have access to limited hardware. It is thus imperative to equip them with latency optimized architectures.

One potential avenue to optimize architectures is through Neural Architecture Search (NAS). In an effort to design better deep learning architectures, NAS provides automated approaches to construct these architectures. Early works based on Reinforcement Learning showed promising results in this area \cite{zoph2016neural}. Although encouraging in their results, early NAS approaches were limited by their massive computational needs. Recent works like DARTS \cite{liu2018darts} and its variants~ \cite{xie2018snas,dong2019search,nayman2019xnas,chen2019progressive,li2019sgas,zhou2019bayesnas,xu2019pc,chu2019fairnas} formulate architecture search as a differentiable problem, which greatly alleviates computational complexity. Further advancements are proposed in recent work like SGAS \cite{li2019sgas}, where optimal architectures show better generalization between search and final tasks. SGAS is also the first work to perform NAS on point cloud architectures. Whatever their underlining processes, these NAS techniques optimize architectures that increase accuracy on a specific task. This approach does not take into account other constraints like latency. For point cloud architectures in hardware bounded applications, it is beneficial to have NAS frameworks that constrain the latency of the output models.

Few works exist in the literature that formulate latency constrained NAS frameworks. Some of these works use a lookup table to obtain the latency of each operation, and approximate the total latency as a sum of all individual latencies \cite{cai2018proxylessnas,wu2019fbnet}. However, these methods are only able to predict latency for chain-style architectures since they ignore the architecture's topology. Other works use more complicated approaches based on reinforcement learning \cite{zoph2016neural}. These approaches are unsuitable for large and complex search spaces \cite{wu2019fbnet} such as directed acyclic graph (DAG) - style architectures. In the DAG space, network latency depends not only on its operations but also on its topology. Therefore, simply aggregating individual latencies is a bad approximation of the total network latency. We alleviate this problem by first introducing a latency regressor that is able to accurately predict a point cloud architecture latency. We then incorporate this regressor as part of a differentiable NAS optimization pipeline. We use a hinge latency loss to constrain the output latency to a desired target. We demonstrate the benefits of our loss formulation compared to other approaches in Section \ref{sec:experiments}. Moreover, our latency aware formulation enables search for point cloud architectures with latencies specific to any desired hardware. 

In this work, we propose a framework to constrain point cloud NAS methods in order to  increase accuracy \emph{and} minimize latency. We achieve this by introducing a differentiable latency estimator, which can be used to define targeted latency constraints in NAS pipelines. Since these constraints are targeted, \ie, we can specify a target latency for the optimal architecture, we can now use NAS solutions to search for the best architecture under a given latency constraint. This effectively allows us to move along the trade-off curve between accuracy and latency to an optimal point for any given application. In this paper, we add our latency constraint to SGAS \cite{li2019sgas}, in order to find latency aware Graph Convolutional Networks (GCNs), that effectively operate  on point clouds. We dub this new framework Latency Constrained Neural Architecture Search (\myNAS). Note that although we use SGAS in this work, our latency constraints can easily be incorporated into most NAS approaches. We show how \myNAS can create point cloud architectures with a range of targeted latencies, while still maximizing accuracy in the end task (Section \ref{sec:modelnet40}). Although the \myNAS architectures are optimized for point cloud classification, we show how they successfully transfer to the task of point cloud part segmentation (Section \ref{sec:partnet}). Such transferability is a testament to the power of our approach.

\mysection{Contributions} We summarize our contributions as three-fold. \textbf{(1)} We propose a novel latency constraint to point cloud NAS pipelines. We show how we can incorporate our constraints in \myNAS, a novel framework on top of SGAS \cite{li2019sgas} to output GCNs for point cloud processing. \myNAS can successfully output architectures with a target latency, while increasing the accuracy of a downstream task. \textbf{(2)} We use \myNAS to search for optimal point cloud classification architectures under a range of target latencies. Our output architectures achieve state-of-the-art results on the ModelNet40 dataset \cite{ds_modelnet}, while significantly reducing latency of the final architecture. \textbf{(3)} We successfully transfer the models learned for classification to the task of part segmentation on PartNet \cite{ds_partnet}, where we also obtain state-of-the-art results in both accuracy and model latency. To the best of our knowledge, we are the first to propose a latency constrained NAS framework for point cloud tasks. 
\section{Related Work}
\mysection{NAS}
Hand-crafted deep learning architectures have achieved considerable success in a wide range of tasks \cite{he2016deep,huang2017densely,krizhevsky2012imagenet,lecun1998gradient,simonyan2014very,GoogLeNet2015}. These innovations were results of human intelligence and experimentation. Initial works based on reinforcement learning attempted to automate the process of architecture search
% The work in 
\cite{zoph2016neural}.
% is the first successful attempts to search for model architecture using reinforcement learning. 
NASNet \cite{zoph2018learning} and ENAS \cite{pham2018efficient} proposed to look for neural architectures in a cell-based search space, and applied regularization and weight sharing techniques to increase search efficiency. PNAS \cite{liu2018progressive} used a sequential model-based optimization (SMBO) strategy to search for structures of increasing complexity. PNAS can reduce computational cost by a  factor of $8$ compared to NASNet. However, it still requires thousands of GPU hours. 
Many recent works \cite{brock2017smash,bender2019understanding,cai2018proxylessnas} aim to reduce the search time by training a single over-parameterized network with inherited/shared weights. 
For example, DARTS \cite{liu2018darts} and its variants \cite{xie2018snas,dong2019search,nayman2019xnas,chen2019progressive,li2019sgas,zhou2019bayesnas,xu2019pc,chu2019fairnas} relaxed the architecture representation to the continuous domain to make the  search differentiable. 
Recenly, NAS has been explored in 3D \cite{li2019sgas,tang2020searching}. In our study, we pick SGAS \cite{li2019sgas} as the baseline method to take advantage of its high efficiency and generalization ability.

\mysection{Resource constrained NAS}
Considering the hardware limitation, NAS can be implemented with extra constraints such as FLOPs and latency. 
SNAS \cite{xie2018snas} included a resource-constrained regularization in their differentiable optimization. It represented the cost of time by the parameter size, number of FLOPs, and memory access cost (MAC).
In ProxylessNAS \cite{cai2018proxylessnas}, FBNet \cite{wu2019fbnet} and Single-Path NAS \cite{stamoulis2019single} approximate the latency of networks as the sum of the latency of every layers by a latency lookup or a learned operation latency predictor. However, these approaches fail to estimate latency for DAG-style architectures.
In MnasNet \cite{tan2019mnasnet}, real-world latency of each sampled model is measured by running it on a single-thread big CPU core of Pixel 1 phones in the training phase.
In contrast to these methods, \myNAS learns an end-to-end differentiable latency regressor for DAG-style architectures, and adopts a loss function with a target latency as a soft constraint. More concurrent works also try to address latency constraints for DAG-style architectures \cite{xu2020latency} or target latency \cite{hu2020tf}.

\mysection{Deep Learning on Point Clouds}
PointNet \cite{pc_qi2017pointnet} provided the first solution for deep learning directly on point clouds. PointNet operates on point cloud chunks, computing point features, and later aggregates them with an order-invariant operation like max pooling. Each point feature is computed using an MLP, without including information from its neighborhood. Because of the nature of this computation, PointNet falls into a group of \textit{Pointwise Networks}. Within this division fall works with more complex aggregation methods, either by looking at more local context \cite{pc_qi2017pointnetpp,pc_engelmann2018} or using complex aggregations through RNNs \cite{pc_huang2018recurrent,pc_ye20183d}. More recent works define convolution operators based on local spatial relationships. These methods fall under the category of \textit{Point Convolution Networks}. They use local structure to define more versatile convolutional kernels, and have proven very successful in pushing the state-of-the-art in point cloud tasks \cite{pc_tatarchenko2018tangent,pc_kpconv,pc_shellnet,pc_pointcnn,pc_xu2018spidercnn,li2018sonet}. A third type of algorithm leverages the power of \textit{Graph Convolutional Networks (GCNs)}. We choose to work with GCNs in this paper given their high versatility in designing different operators, and their proven results in previous NAS approaches \cite{li2019sgas}. Following is a brief survey of GCNs and their applications.

\mysection{Graph Convolutional Networks (GCNs)}
Recent years have seen a surge of non-Euclidean data in real-world scenarios. Such representation is prime for GCNs, where convolution operators are designed to work with generic graph representations of data. Several such GCNs exist in the literature for a wide number of applications \cite{kipf2016semi,hamilton2017inductive,velivckovic2017graph,pham2017column,xu2020gtad}. In the area of point clouds, DGCNN \cite{dgcnn} introduced EdgeConv, and used it to conduct dynamic graph convolution on point clouds. Recently, DeepGCNs \cite{Li2019DeepGCNs,Li2019DeepGCNsMG} integrated residual/dense connections and dilated convolutions into GCNs for point clouds. These integrations enabled them to successfully train GCNs of up to 112 layers. The operations in DGCNN and DeepGCNs were used in SGAS to search for optimal GCNs, and we similarly apply them in \myNAS for our latency constrained architectures.

\begin{figure}
    \centering
    \includegraphics[width=1\textwidth,trim={2.5cm 3cm 4.5cm 2cm}] {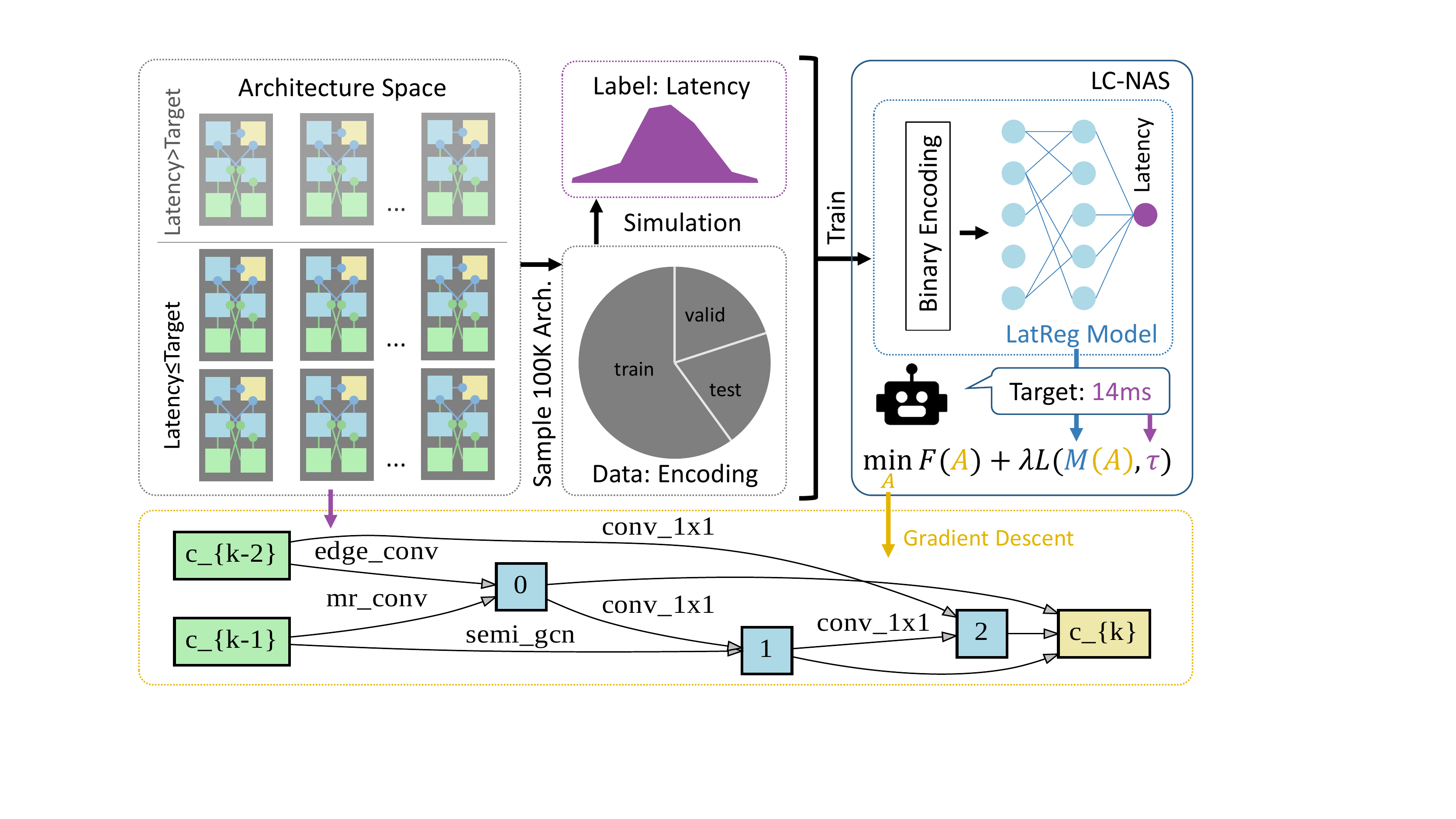}
    \caption{\textbf{\myNAS pipeline.} We sample a large number of architecture from our defined search space and measure their latencies. We use these architectures to train a latency predictor. Our predictor takes as input a binary encoding of the architecture and provides a latency measure. We integrate this predictor as a constrain into the NAS loss, where the target latency is part of the loss. We optimize our pipeline to discover point cloud architectures that meet a target latency. With this targeted approach, we can easily trade-off between accuracy and latency, and discovery models specific to a given deployment architecture.}
    \label{fig:pipeline}
\end{figure}
\section{Methodology}

\subsection{Preliminary - Sequential Greedy Architecture Search}
In order to calibrate mismatching rankings between architectures during the search phase and the evaluation phase, \emph{Sequential Greedy Architecture Search} (SGAS)~\cite{li2019sgas} proposes to solve the bi-level optimization in DARTS in a sequential greedy fashion. Following DARTS, SGAS searches for cells and composes networks by stacking cells with identical structure. A cell is typically represented as a directed acyclic graph (DAG) with $N$ nodes including two input nodes, several intermediate nodes and one output node. The $i$-th topological ordered node in the DAG is an intermediate feature representation denoted as $x^{(i)}$. Each directed edge $(i, j)$ in the DAG represents an operation $o^{(i, j)}$ that transforms $x^{(i)}$. Each intermediate node $x^{(j)}$ is computed by summing up the outputs of all the edges from its predecessors,
$
	x^{(j)} = \sum_{i<j} o^{(i, j)}(x^{(i)})
$.
By parameterizing operation $o^{(i, j)}$ with architectural parameters $\alpha^{(i, j)}$ over the space of all the candidate operations $\mathcal{O}$, DARTS turns the architecture search into a differentiable form. During the search phase, the operation $o^{(i, j)}$ is relaxed as a mixture operation,
$
	\bar{o}^{(i,j)}(x^{(i)}) = \sum_{o \in \mathcal{O}} \frac{\exp(\alpha_o^{(i,j)})}{\sum_{o' \in \mathcal{O}} \exp(\alpha_{o'}^{(i,j)})} o(x^{(i)})
$.
After the search, the discrete architecture is obtained by choosing a non-zero operation with the highest weight for every mixture operations,
$
	o^{(i,j)} = 
	\mathrm{argmax}_{o \in \mathcal{O}} \enskip \alpha^{(i,j)}_o
$.
However, the discrepancy between the search and evaluation phases and the negative effect of weight sharing in DARTS make the searched architectures fail to generalize in the final evaluation. In order to alleviate these problems, SGAS proposed to determine and prune operations progressively during the search phase in a greedy fashion. Theoretically, we can use any differentiable methods such as DARTS \cite{liu2018darts} or its variants \cite{chen2019progressive,xie2018snas,xu2019pc}. However, we opt to choose SGAS as our base method (as apposed to DARTS for example) for its efficiency and more stable results.

\subsection{Latency Constrained Neural Architecture Search}
The goal of our work is to automatically design a well-performing GCN architecture that is able to run on a specific hardware platform within a given target latency. To achieve this goal, we need to take the latency constraint into consideration in the optimization during the search phase. As mentioned, previous works either build a lookup table for the latency of each operations and approximate the latency by summing up corresponding latency sequentially, or use reinforcement learning to optimize the latency by treating it as a part of the score. However, none of them is feasible to apply to differential NAS methods with a search space as a DAG, since the latency of a network with a general DAG structure depends on its topology and can not be estimated by simply summing up the latency of each operations. To this end, we propose to first learn a latency regressor that is able to predict the latency of all the possible architectures on a specific device within the search space. And then we incorporate the learned latency regressor as a constraint in the bi-level optimization objective and regularize the architectural parameters $\mathcal{A} = \{\alpha_o^{(i,j)}~|~\forall (i,j)\}$. 

% 
% \subsection{Search with a Latency Constraint}

\mysection{Search Space}
We use the same search space as SGAS. Each edge in our graph convolutional cell has $10$ candidate operations including \textit{Skip-Connect}, \textit{Conv-1$\times$1}, \textit{EdgeConv} \cite{dgcnn}, \textit{MRConv} \cite{Li2019DeepGCNs}, \textit{GAT} \cite{velivckovic2017graph}, \textit{SemiGCN} \cite{kipf2016semi}, \textit{GIN} \cite{Xu2018HowPAGIN}, \textit{SAGE} \cite{hamilton2017inductive}, \textit{RelSAGE} \cite{li2019sgas} and \textit{Zero}. Please refer to the original SGAS paper \cite{li2019sgas} for more details of these GCN operators. K Nearest Neighbours (KNN) is used to take the input vertex/point features of a cell to build a KNN graph by constructing dynamic edges. Obtained edges are then shared with others operations within the cell. Following DeepGCNs \cite{Li2019DeepGCNs}, dilated graph convolutions with a linearly increasing dilation rate is used in cells. In our experiment, each cell contains $N = 6$ nodes with $3$ intermediate nodes and $9$ edges during the search phase. For each intermediate node in the DAG, we retain $2$ incoming edges after the search process. Therefore, we will obtain a DAG with $6$ edges at the end of the search process. There are about $18$ million possible architectures in this search space. It is impossible to enumerate or measure the latency of these millions of architectures. Therefore, we propose to train a latency approximator from the data itself. To learn a precise latency regressor, we need to first sample sufficient architectures from the search space and measure their latency to create a dataset. Then we can train a regression network that takes an architecture encoding as input and output the predicted latency.

\mysection{Architecture Encoding} 
To encode the model architectures, we need to encode the graph convolutional cell, the basic building block in cell-based neural architecture search. Our search space is a DAG with $9$ edges and $10$ candidate operations for each edge. After the search, $6$ edges are retained in total. Thus, we can use a $9\times 10$ binary encoding matrix $\mathbf{E}\in \{0,1\}^{9\times10}$ to represent the cell, where $e_{m, n}=1$ indicates that the operation $n$ is chosen for the edge $m$. Since we only retain $6$ edges, the encoding matrix $\mathbf{E}$ a sparse matrix with exactly $6$ entries with values of $1$.

\mysection{Latency Regressor} \label{sec:lat_reg}
In this work, we consider point cloud classification on ModelNet$40$ as our target task and NVIDIA RTX $2080$ as our target hardware platform. We sample and measure $100K$ random cell-architectures from the search space. We stack the sampled cell $3$ times with a channel size of $128$. For more details about the model hyper-parameters, please refer to the experiment \secLabel \ref{sec:modelnet40}. The latency is measured by the inference time of the sampled architectures with randomly initialized weights and a random tensor of shape $(\textit{batch size} = 1, \textit{feature dims} = 3, \textit{num of points} = 1024)$. The data is then split into three folds with $60\%$ as the training subset, $20\%$ as the validation subset, and $20\%$ as the testing subset.
The latency ranges from $5.9$ to $23.5$ milliseconds (ms).
We leverage a Multi Layer Perceptron (MLP) as our latency regressor (LatReg). Given the encoding matrix $\mathbf{E}$, we first vectorize $\mathbf{E}$ and feed it into three fully-connected layers. We respectively set $256$, $128$ and $1$ neurons in each of the three layers, interleaved with sigmoid activation functions. Finally, the MLP produces one scalar value as the latency prediction. 
During training, we normalize the latency by the mean $\mu_{train}=15.32$ms and the standard deviation $\sigma_{train}=2.24$ms. We train the network from scratch, and set the batch size to $256$. We employ Adam optimizer in PyTorch with the default parameters: learning rate equal to $0.001$ and betas equal to ($0.9$, $0.999$). We use mean square error (MSE) as the loss function. From the loss curve of the validation set, we find the model saturated efficiently on the $70$-th epoch. The LatReg model reaches an average absolute error of $0.16$ms on the test subset.

\mysection{Loss function with a Target Latency as Constraint} The learning procedure of neural architecture search can be formulated as a bi-level optimization problem \cite{liu2018darts}. Previous resource-aware differentiable NAS methods usually add/multiply the latency loss as a regularizer to the cross-entropy loss \cite{cai2018proxylessnas,wu2019fbnet,xie2018snas}. However, these methods are not able to minimize the architecture to be lower than a certain latency, which is considered to be very important for the deployment on a specific hardware platform. Moreover, the regularization weight is hard to tune. A big regularization weight leads to efficient/fast models but with low capacity. On the other end. a small regularization weight fails to obtain efficient models. Therefore, we propose to use a hinge-loss-like regularization loss for the latency constraint as follows: 
\begin{align}
	\min_\mathcal{A} \quad & \mathcal{L}_{val}(\mathcal{W}^*(\mathcal{A}), \mathcal{A}) {~\color{blue} + \lambda\max( LatReg(\mathcal{E}(\mathcal{A})) - target, ~0) } \label{eq:outer} \\
	\text{s.t.} \quad &\mathcal{W}^*(\mathcal{A}) = \mathrm{argmin}_\mathcal{W} \enskip \mathcal{L}_{train}(\mathcal{W}, \mathcal{A}) \label{eq:inner}
\end{align}
where $\mathcal{L}_{val}$ is the cross-entropy loss on validation set, $\mathcal{L}_{train}$ is the cross-entropy loss on training set, $\mathcal{A}$ is the architectural parameters, $\mathcal{W}$ is the network weights, $\lambda$ is the regularization factor, $LatReg(\cdot)$ is the learned latency regressor, $target$ is the target latency and $\mathcal{E}(\cdot)$ is a non-differentiable binarized function that take as input continuous architectural parameters $\mathcal{A}$ and output a binarized architecture encoding $\mathbf{\tilde{E}}$. The advantages of this hinge latency loss are as follow: (1) Once the predicted latency is lower than the target latency, the latency loss term is zero and the bi-level optimization reduces into the original objective function. This reduces the risk of harming the model capacity. (2) For the same reason, the regularization factor $\lambda$ becomes less sensitive and easy to tune. (3) Incorporating a target latency into the latency loss makes the architecture search more controllable and deployable on a hardware of desired specific latency. Binarizing the continuous architectural parameters $\mathcal{A}$ before predicting the latency is necessary since $LatReg(\cdot)$ is trained on binary inputs. If we use $\mathcal{A}$ as input, the latency prediction of $LatReg(\cdot)$ would in inaccurate due to the discrepancy between continuous architectural parameters and the binary architecture encoding. However, the binarized function $\mathcal{E}(\cdot)$ is non-differentiable since it involves some rules/heuristics to derive a discrete architecture encoding such as choosing $6$ edges out of $9$ and selecting a non-zero operation with the highest weight. In order to obtain the gradient of the latency loss with respect to architectural parameters $\mathcal{A}$, we introduce an approximated gradient-based approach.

\mysection{Optimizing the Latency Constraint}
As mentioned, the binarized function $\mathcal{E}(\cdot)$ is non-differentiable. It mainly includes a softmax operation over architectural parameters, choosing two edges for each intermediate nodes based on pre-defined rules/heuristics and selecting a non-zero operation with the highest weight for the corresponding edge. The parts of choosing edges and operations make $\mathcal{E}(\cdot)$ non-differentiable. One conventional way to update the real-valued weight $\mathcal{A}$ is to use the gradient with respect to its binarized value $\mathbf{\tilde{E}}$ which is proposed in BinaryConnect \cite{courbariaux2015binaryconnect} and has also been used ProxylessNAS \cite{cai2018proxylessnas}. However, this approach does not take real-valued weight into consideration while computing the gradient. We propose a modified approximated gradient-based approach to optimize the architectural parameters. We denote the softmax output of $\beta_{m, n} = softmax(\alpha_{m, n}|\bm{\alpha}_{m}) = \frac{\exp(\alpha_{m, n})}{\sum_{k}\exp(\alpha_{m, k})}$, where $\alpha_{m,n}$ is the architectural parameter of operation $n$ of edge $m$. To compute the gradient of $\mathcal{E}(\cdot)$ with respect to $\mathcal{A}$, we trust the selection rules/heuristics as a linear operation by approximating with multiplying an element-wise mask $\bm{\zeta}$, where $\zeta_{m,n} = \frac{1}{\beta_{m,n}}$ if $n = n^{*}$ and $\zeta_{m,n} = 0$ if $n \neq n^{*}$. Note that $n^{*}$ is the chosen operation of edge $m$. Therefore the binarized function becomes $\mathcal{E}(\alpha_{m,n}) = \tilde{e}_{m,n} \approx \beta_{m, n} \cdot \zeta_{m,n}$ and the gradient can be obtained. We denote the latency loss term as $\mathcal{L}_{lat}$. We have:
\begin{align}
	\frac{\partial \mathcal{L}_{lat}}{\partial \alpha_{m,n}} 
	= \sum_{k} \frac{\partial \mathcal{L}_{lat}}{\partial \beta_{m,k}} \cdot \frac{\partial \beta_{m,k}}{\partial \alpha_{m,n}}
	= \sum_{k} \frac{\partial \mathcal{L}_{lat}}{\partial \tilde{e}_{m,k}} \cdot \frac{\partial \tilde{e}_{m,k}}{\partial \beta_{m,k}} \cdot \frac{\partial \beta_{m,k}}{\partial \alpha_{m,n}}
	\label{eq:gradient1}
\end{align}
where $\frac{\partial \beta_{m,k}}{\partial \alpha_{m,n}} = \beta_{m,n} - \beta_{m,n}^{2}$ if $n = k$ and $\frac{\partial \beta_{m,k}}{\partial \alpha_{m,n}} = - \beta_{m,n} \cdot \beta_{m,k}$ if $n \neq k$. Since $\frac{\partial \tilde{e}_{m,k}}{\partial \beta{m,k}} = \zeta_{m,k}$. We obtain the gradient as follows:
\begin{equation*}
\frac{\partial \mathcal{L}_{lat}}{\partial \alpha_{m,n}}
= \frac{\partial \mathcal{L}_{lat}}{\partial \tilde{e}_{m,n^{*}}} \cdot \frac{1}{\beta_{m,n^{*}}} \cdot \frac{\partial \beta_{m,n^{*}}}{\partial \alpha_{m,n}}
= \begin{cases}
\frac{\partial \mathcal{L}_{lat}}{\partial \tilde{e}_{m,n^{*}}} \cdot (1 - \beta_{m,n^{*}}) &\text{for $n = n^{*}$}\\
\frac{\partial \mathcal{L}_{lat}}{\partial \tilde{e}_{m,n^{*}}} \cdot - \beta_{m,n}  &\text{for $n \neq n^{*}$}
\end{cases}
\end{equation*}
In this way, we can update the architectural parameters $\mathcal{A}$ using the gradient above. Therefore, the latency constraint can be optimized during search.
\section{Experiments}
\label{sec:experiments}
As mentioned in \secLabel \ref{sec:lat_reg}, our target task is classification on ModelNet40 using NVIDIA RTX 2080. We sample $100 K$ architectures from the search space and measure their latency  to build a dataset. Then, we train a latency regressor on the latency dataset. After that, we use the pre-trained latency regressor to constrain the architecture search on ModelNet$10$ with latency targets on ModelNet$40$ ranging from $6$ms to $18$ms. We then evaluate the performance of obtained architectures on ModelNet$40$ by training from scratch. We also transfer our architectures to a completely different task, part segmentation, to show the generalization of the architectures. Finally, we conduct a thorough ablation study to demonstrate the effects of regularization strength, loss function, and the choice of hyper-parameters for our models.

\subsection{\myNAS on ModelNet$10$} 
\mysection{Dataset} ModelNet \cite{ds_modelnet} is a classical dataset for 3D point cloud classification, which has two subsets ModelNet$10$ and ModelNet$40$ containing objects with $10$ and $40$ classes respectively.
ModelNet$10$ comprises $4,899$ 3D models split into $10$ classes, with $3,991$ models in training and 908 models in testing. ModelNet$40$ consists of 12,311 models split into $40$ classes, with 9,843 models in training and $2,468$ in testing. The goal of classification on ModelNet datasets is to classify the category of a 3D model. We first search GCN architectures on ModelNet$10$ using \myNAS and evaluate their performance on ModelNet$40$.

\mysection{Training Settings}
During the training of the search phase, $1024$ points with only 3D coordinates $(x, y, z)$ are sampled from the 3D models in ModelNet$10$ as input.
We set the regularization factor $\lambda$ as $0.5$ and vary the target latency from $6$ms to $18$ms with a step of $2$ms. The other settings follow those of SGAS. We use $2$ cells with $32$ initial channels and search for the architectures for $50$ epochs with batch size $28$. Two different optimizers are used for optimizing model weights $\mathcal{W}$ and $\mathcal{A}$. We use SGD with initial learning rate $0.005$, momentum $0.9$, and weight decay $3\times10^{-4}$ to optimize the model weights. An Adam optimizer with an initial learning rate $3 \times 10^{-4}$, momentum $(0.5, 0.999)$, and weight decay $10^{-3}$ is used for architecture parameters $\mathcal{A}$. We use the same edge selection criterion as SGAS \textit{Criterion 2}. \myNAS begins to determine one operation for a selected edge every $7$ epochs after warming up for $9$ epochs. A history window of size $4$ is used for \textit{Selection Stability}. We increase the batch size  by $4$ after each decision. The total time for a search run is around $0.19$ GPU days on a single NVIDIA GTX 1080Ti, which is the same for SGAS. This means the extra computation overhead of adding the latency constraint is negligible.

\subsection{Architecture Evaluation on ModelNet$40$} \label{sec:modelnet40}
After searching on ModelNet$10$, we get $7$ searched cell structures with target latency $6$ms, $8$ms, $10$ms, $12$ms, $14$ms, $16$ms and $18$ms respectively. We build a large network for each cell with the same hyper-parameters that are used to generate the architecture for learning the latency regressor. We then train them on ModelNet40 from scratch. We evaluate the performance on ModelNet$40$ using two metrics: the overall accuracy (O.A.) and class accuracy (Class ACC.).

\mysection{Training Settings}
$1024$ points with 3D coordinates are used as input. We stack the searched cell $3$ times with channel size $128$. An MLP with $1024$ neurons is used to fuse the concatenation of all the output features of $3$ cells. And then, the fused features are aggregated through a max-pooling layer and an average pooling layer. We concatenate the aggregated features from two pooling layers and feed them into a 3-layer MLP classifier with $\{512, 256, 40\}$ neurons to classify the input point clouds into 40 categories. For the first two MLP layers, we use dropout layers with probability $0.5$ during training. A drop path is applied with probability $0.2$. SGD is used to optimize the model weights with initial learning rate $0.1$, momentum $0.9$, and weight decay $1\times10^{-4}$. We use a cosine annealing learning rate scheduler with a minimum learning rate of $0.001$. Our architectures are all trained for $400$ epochs with batch size $32$.

\mysection{Evaluation Results and Analysis}
We report the best overall accuracy (O.A.) and the corresponding class accuracy (Class ACC.) on the test dataset for all the $7$ discovered architectures in \tblLabel \ref{tab:modelnet10}.
We observe in \tblLabel \ref{tab:modelnet10} that \myNAS is able to meet the target latency in the vast majority of cases (only exception is \myNAS-10, where the actual latency is only over target by $1$ms). We also show strong accuracy results, and a meaningful trend of dropping accuracy as target latency increases. We also compare our discovered architectures with state-of-the-art results in \tblLabel \ref{tab:modelnet40sota}. We see how \myNAS can obtain low latency and still maintain competitive results in terms of accuracy. Particularly, we observe how we drop the baseline latency (SGAS in \tblLabel \ref{tab:modelnet40sota}) from $16.62$ms to $5.47$ms with an accuracy drop of less than $3$ points. In a real-world scenario, where such architectures need to be deployed in hardware-bounded applications, \myNAS provides architectures suitable for latency bound scenarios.

\begin{table}[ht]
    % center the remaining to the page
    \centering
    \small
    % resize box to fit within the page width
    % Captioning
    \caption{\textbf{Evaluation Results on ModelNet40.} We show results obtained on ModelNet40 using architectures discovered on different target latencies. We report the target latency, predicted latency by our latency regressor, actual measured latency, and number of parameters. We also show overall accuracy and class accuracy. We observe that our architectures consistently meet the target constraint while achieving high accuracy.}
\begin{tabular}{l||c|c|c||c|c|c}
  & \multicolumn{3}{c||}{\B Latency (ms)} & & \multicolumn{2}{c}{\B Accuracy} \\
\B Method & \B Target & \B Predicted & \B Measured & \B \# Param. & \B Overall  & \B Class \\\hline\hline
% 20 & 14.22 & 13.63 & 3.49 & 3.90 & 92.54 & 89.98 \\\midrule
\B \myNAS-18  & 18 & 17.06 & 16.66 & 3.91 & 92.79 & 89.66 \\\hline
\B \myNAS-16  & 16 & 13.71 & 13.57 & 3.91 & 92.62 & 90.13 \\\hline
\B \myNAS-14  & 14 & 12.64 & 12.41 & 3.91 & 92.42 & 89.16 \\\hline
\B \myNAS-12  & 12 & 10.07 & 9.96  & 3.85 & 92.34 & 89.57 \\\hline
\B \myNAS-10  & 10 & 11.02 & 11.09 & 3.86 & 92.75 & 90.76 \\\hline
\B \myNAS-8   &  8 & 7.84  & 7.51  & 3.71 & 90.40 & 85.36 \\\hline
\B \myNAS-6   &  6 & 6.12  & 5.47  & 3.61 & 90.51 & 84.71 \\\hline \hline
\B  Average   &  - & 11.21 & 10.95 & 3.82 & 91.98 & 88.48 \\\midrule

\end{tabular}
    % Labeling
    \label{tab:modelnet10}
\end{table}

\begin{figure}[t]
    \centering
    \includegraphics[width=1\textwidth,trim={0.5cm 8.3cm 0.5cm 3cm},clip]{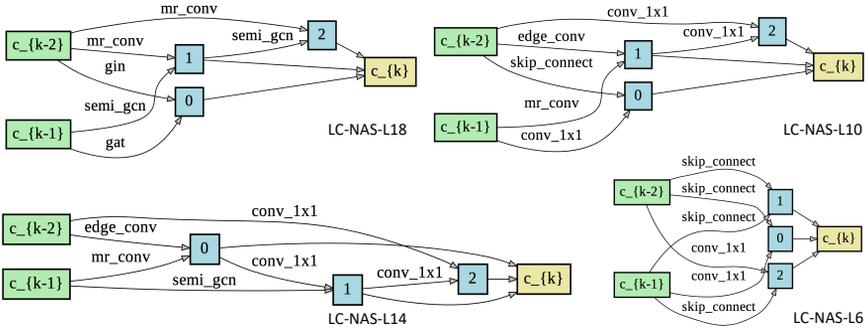}
    \caption{\textbf{Visualization of discovered architectures.} We show the cell operations of 4 architectures discovered with 4 target latencies, $6$ms, $10$ms, $14$ms, and $18$ms (denoted by \myNAS-6, \myNAS-10, \myNAS-14, and \myNAS-18 respectively). We observe how the operations increasing in complexity as the target latency increases. Low latency architectures favor $1 \times 1$ convolutions and skip connections, while the larger ones go for complex GCN operations like \textit{MRConv} and \textit{EdgeConv}.}
    \label{fig:visualizations_gcn}
\end{figure}

\vspace{-20pt}
\begin{table}[ht]
    % center the remaining to the page
    \centering
    \small
    % resize box to fit within the page width
    % Captioning
    \caption{\textbf{Comparison to state-of-the-art methods. } We measure the latencies of state-of-the-art architectures on ModelNet40 with their reported accuracies. SGAS is our baseline since it is equivalent to our approach without any constraints. We show the significant latency reduction of SGAS while maintaining comparable accuracies.}
\begin{tabular}{l|r|c|c|l|r|c}
\B  Method   & \B Latency &   \B Overall Acc. & ~~~& \B Method   & \B Latency &   \B Overall Acc. \\\hline\hline

\B PointNet \cite{pc_qi2017pointnet}   & 4.21  ms &   89.2 & ~~~& \B \myNAS-18  & 16.66 ms  & 92.79 \\
\B PointNet++ \cite{pc_qi2017pointnetpp} & 23.51    ms &   90.7 & ~~~& \B \myNAS-16  & 13.57 ms  & 92.62 \\
\B DGCNN \cite{dgcnn}   & 9.42  ms &   92.2 & ~~~& \B \myNAS-14  & 12.41 ms  & 92.42 \\
\B PointCNN \cite{pc_pointcnn}  & 26.79  ms &   92.2 & ~~~& \B \myNAS-12  &  9.96 ms  & 92.34 \\
\B ShellNet \cite{pc_shellnet}  & 9.29	    ms & \B93.1 & ~~~& \B \myNAS-10  & 11.09 ms  & 92.75 \\
\B KPConv	\cite{pc_kpconv}   & 26.81	    ms &   92.9 & ~~~& \B \myNAS-8   &  7.51 ms  & 90.40 \\
\B SGAS	 \cite{li2019sgas}    & 16.62	    ms & 92.9 & ~~~& \B \myNAS-6   &  5.47 ms  & 90.51 \\
\B RS-CNN$^*$ \cite{liu2019rscnn} & -  & 92.4 \\ % 92.4 for single RS-CNN

\hline
\end{tabular}
\\
\leftline{~~~$^*$We report the single vote performance for fair comparison.}
    % Labeling
    \label{tab:modelnet40sota}
\end{table}
We visualize in \figLabel \ref{fig:visualizations_gcn} samples from the resulting discovered architectures. As expected, we observe an increase in the complexity of operations as the target latency increases. For example, with target latency of $6$ms (\myNAS-6), we observe that the majority of operations are $1 \times 1$ convolutions and skip connections to satisfy such a low latency budget. As we increase the latency to $18$ms, the observed operations become complex and time-consuming like \textit{MRConv}, \textit{GAT}, and \textit{GIN}. Architectures in between mix between simple and complex operations in order to meet the target latency. The results are visual validations that \myNAS discovers meaningful architectures in line with the required constraints.

\subsection{Transferring Searched Architectures to Part Segmentation} \label{sec:partnet}

The \myNAS architectures searched on ModelNet10 are transferred for the task of Part Segmentation, in particular on the PartNet benchmark~\cite{ds_partnet}.

\mysection{Dataset} 
PartNet is composed of 24 object classes with 1, 2 or 3 difficulty levels for each classes. We focus exclusively on the 17 object classes with the highest difficulty level $3$. Each model takes point clouds of $10000$ points.

\mysection{Training Settings}
We train a model for each object class, as common practice~\cite{ds_partnet}, since each object has a different number of parts. We adapted the PyTorch implementation from \cite{Li2019DeepGCNsMG}. We train each model for $500$ epochs with an initial learning rate of $0.005$ and a decay factor of $0.9$ every $50$ epochs. We use the PyTorch Adam optimizer with default values.

\mysection{Evaluation Results and Analysis}
We report the performances of our \myNAS architectures (Part mIOU) and compare with related works in Table~\ref{tab:partnet}. \myNAS achieves state-of-the-art performance on $12$ classes, and enjoys an average part mIOU of $48.55\%$. We outperform the baseline PointCNN~\cite{pc_pointcnn} by up to $2.06\%$ with $\sim10\times$ speedup. PointCNN is not able to scale efficiently due to the numerous KNN operations. SGAS~\cite{li2019sgas} employs a similar network search strategy but is less time-efficient for similar performances. Our method also outperforms PointNet~\cite{pc_qi2017pointnet} ($35.6\%$) and PointNet++~\cite{pc_qi2017pointnetpp} ($42.5\%$) by a consistent margin.

\begin{table}[ht]
    % center the remaining to the page
    \centering
    % resize box to fit within the page width
    % Captioning
    \caption{
    \textbf{Part Segmentation on PartNet (part mIOU on level 3).}
    Best results in \textbf{bold}.
    \myNAS models provide state-of-the-art performances on $12$ object classes with faster inference time.
    Runtimes are averaged on $1000$ runs with $10000$ input points. 
    }
    \resizebox{\textwidth}{!}{
\begin{tabular}{l||r||c||c|c|c|c|c|c|c|c|c|c|c|c|c|c|c|c|c}
\B Method    &\B(ms) &\B Avg. &\B Bed &\B Bott &\B Chair &\B Clock &\B Dish &\B Disp &\B Door &\B Ear &\B Fauc &\B Knife &\B Lamp &\B Micro &\B Frid &\B Stora &\B Table &\B Trash &\B Vase\\\midrule\midrule
\B PointCNN\cite{pc_pointcnn} 
&  1402 &  46.49 &  41.9 &  41.8 &\B43.9 &  36.3 &  58.7 &  82.5 &  37.8 &  48.9 &\B60.5 &  34.1 &  20.1 &\B58.2 &  42.9 &  49.4 &  21.3 &  53.1 &\B58.9\\\midrule
\B SGAS\cite{li2019sgas}
&  185  &  48.28 &\B43.4 &  50.8 &  41.2 &  38.8 &  61.4 &  82.6 &  37.1 &  48.8 &  56.1 &  49.4 &  21.2 &  56.5 &  44.5 &\B49.4 &  29.3 &  54.4 &  56.0\\\midrule
\B deep LPN\cite{le2020going}
& 191  &  38.60 &  29.5 &  42.1 &  41.8 &  34.7 &  33.2 &  81.6 &  34.8 &\B49.6 &  53.0 &  44.8 &\B28.4 &  33.5 &  32.3 &  41.1 &\B36.3 &  43.1 &  57.8\\\midrule\midrule
\B\myNAS-10 
&\B143  &  48.10 &  41.4 &  50.5 &  39.6 &  37.8 &  61.1 &\B82.9 &  37.4 &  48.4 &  53.6 &  48.5 &  22.3 &  57.8 &\B46.6 &  47.9 &  31.1 &\B54.8 &  56.0\\\midrule
\B\myNAS-14 
&  152  &\B48.55 &  41.9 &\B51.7 &  39.7 &\B39.6 &\B61.5 &  82.5 &\B39.3 &  49.0 &  54.7 &\B55.3 &  22.2 &  55.1 &  45.2 &  48.0 &  30.3 &  54.6 &  54.9\\\midrule
\B\myNAS-18 
&  185  &  46.60 &  40.7 &  50.5 &  39.9 &  39.5 &  59.8 &  82.2 &  35.0 &  44.5 &  53.2 &  44.9 &  22.0 &  54.1 &  41.5 &  45.8 &  31.5 &  53.0 &  54.4\\\midrule
\end{tabular}
    }
    
    % Labeling
    \label{tab:partnet}
    \vspace{-10pt}
\end{table}

\mysection{Timing}
Our \myNAS architecture targets a specific latency with $1024$ points for inputs, but PartNet provides object point clouds with $10000$ points each.
Our implementation of KNN for PointCNN~\cite{pc_pointcnn}, SGAS~\cite{li2019sgas} and \myNAS is not optimized hence suffers the large input size scale up.
As an example, the latency of PointCNN jumps from $26$ms for $1024$ points to $1402$ms for $10000$ points on a NVIDIA GTX2080 with $8$GB.
Similarly, our models searched for a latency of $10$, $14$ and $18$ms jumps to $143$, $152$ and $185$ms for those larger point clouds.
All times reported are averaged over $1000$ runs after $50$ runs of warm-up (for Pytorch to optimize the CUDNN operations), using one batch of $10000$ points.

\subsection{Ablation Study on ModelNet$40$}
\mysection{Ablation on Latency Constraint Loss} In \tblLabel \ref{tab:ablation1}, we conduct an ablation study on our hinge latency loss by experimenting with different regularization factors $\lambda = 0.8$ and $\lambda = 1$ with the same search setting as $\lambda = 0.5$ shown in \tblLabel \ref{tab:modelnet10}. \textit{Note that the results shown in \tblLabel \ref{tab:modelnet10} are trained with the hinge latency loss defined in \eqLabel \ref{eq:outer} with $\lambda = 0.5$. The average latency, overall accuracy and class accuracy are $10.95$ms, $91.98\%$ and $88.48\%$ respectively when $\lambda = 0.5$.} Compared with $\lambda = 0.8$ and $\lambda = 1$, the differences of accuracy and latency are within $0.1\%$ and $0.4$ms. Therefore, it is clear that the proposed hinge latency loss function is fairly robust to different strengths of regularization. Furthermore, we perform an experiment by replacing the hinge latency loss with a mean square error (MSE) loss with regularization factor as $\lambda = 0.5$. Architectures searched with MSE loss as the constraint only perform $0.04\%$ better than the hinge latency loss counterpart but have on average $1.08$ms more latency. This shows that searching the hinge latency loss obtains more efficient architectures than searching MSE loss.

\mysection{Ablation on Hyper-parameters of Models}
In \tblLabel \ref{tab:ablation2}, we conduct an ablation study of searched cells with different numbers of cells and different input channels size. It shows that the obtained cells are scalable with different hyper-parameters. Architectures with more cells and a larger channel size usually perform better. For example, architectures with 2 cells and 128 channels outperform those with 1 cells and 128 channels by $0.26\%$ in term of overall accuracy. Small architectures can still achieve state-of-the-art results. LC-NAS-18 with 1 cell and 64 channels reaches $92.30\%$ O.A. with only $7.39$ms inference time.

\begin{table}[t]
    % center the remaining to the page
    \centering
    \small
    % resize box to fit within the page width
    % Captioning
    \caption{\textbf{Ablation on Latency Constraint Loss on ModelNet40.} We show the effect of different $\lambda$ parameters in our constrained hinge loss. \textit{We compare these results with \tblLabel \ref{tab:modelnet10}, where $\lambda=0.5$ and the average latency, overall accuracy, and class accuracy are $10.95$ms, $91.98\%$ and $88.48\%$ respectively.} When tweaking $\lambda$ to $0.8$ and $1.0$, the results change slightly. This shows our hinge loss is robust to different regularization strengths. We also change the hinge loss to MSE and observe a significant increment in average latency. Our hinge loss is better at predicting tighter latency networks.}
\begin{tabular}{l||r|c|c||r|c|c||r|c|c}
\multirow{2}{*}{\B Method} & \multicolumn{3}{c||}{\B $\mathbf{\lambda=0.8}$}&\multicolumn{3}{c||}{\B $\mathbf{\lambda=1}$}& \multicolumn{3}{c}{\B MSE loss}\\\cline{2-10}
 & \B Latency & \B O.A. & \B C.A.& \B Latency & \B O.A. & \B C.A.& \B Latency & \B O.A. & \B C.A.\\\hline
%20 & 14.73 & 92.62 & 90.17 & 12.66 & 92.46 & 88.71 & 20.64 & 92.54 & 89.39 \\ 
\B \myNAS-18 & 16.61 ms & 92.83 & 90.31 & 18.35 ms & 92.87 & 90.02 & 17.34 ms & 92.54 & 90.38 \\
\B \myNAS-16 & 12.69 ms & 92.38 & 89.24 & 14.41 ms & 92.54 & 89.50 & 14.44 ms & 92.59 & 89.83 \\
\B \myNAS-14 & 13.20 ms & 92.71 & 89.88 & 12.33 ms & 92.63 & 90.13 & 13.41 ms & 91.53 & 86.80 \\
\B \myNAS-12 & 11.05 ms & 92.54 & 90.94 & 10.69 ms & 92.91 & 89.45 & 14.22 ms & 92.34 & 89.09 \\
\B \myNAS-10 & 11.34 ms & 92.30 & 89.95 & 8.29  ms & 90.88 & 85.15 & 10.58 ms & 92.71 & 89.61 \\
\B \myNAS-8  & 7.67  ms & 90.03 & 84.30 & 6.42  ms & 90.36 & 84.84 & 7.87  ms & 92.13 & 88.13 \\
\B \myNAS-6  & 6.61  ms & 90.96 & 85.71 & 5.86  ms & 90.96 & 85.71 & 6.36  ms & 90.32 & 85.08 \\\hline
\B  Avg. & 11.31 ms & 91.96 & 88.62 & 10.91 ms & 91.88 & 87.83 & 12.03 ms & 92.02 & 88.42 \\\hline
\end{tabular}
    
    % Labeling
    \label{tab:ablation1}
\end{table}

\begin{table}[t]
    % center the remaining to the page
    \centering
    \small
    % resize box to fit within the page width
    % Captioning
    \caption{\textbf{Ablation on Hyper-parameters of Models.} We change the number of channels in the operations and the total number of cells used in the networks and evaluation the architectures on ModelNet40 .}
\begin{tabular}{l||c|c|c||c|c|c||c|c|c}
\multirow{2}{*}{\B Method}  & \multicolumn{3}{c||}{\B 1 cell, 64 channels}&\multicolumn{3}{c||}{\B 1 cell, 128 channels}& \multicolumn{3}{c}{\B 2 cells, 128 channels}\\\cline{2-10}
 & \B Latency & \B O.A. & \B C.A.& \B Latency & \B O.A. & \B C.A.& \B Latency & \B O.A. & \B C.A.\\\hline
\B \myNAS-18 & 7.39 ms & 92.30 & 88.13 & 6.36 ms & 92.22 & 89.09 &11.32 ms   &92.26    & 88.66  \\
\B \myNAS-16 & 4.85 ms & 91.77 & 87.85 & 5.14 ms & 92.15 & 89.18 &10.90 ms   &92.59    & 89.63  \\
\B \myNAS-14 & 4.85 ms & 92.02 & 88.04 & 5.58 ms & 92.30 & 89.55 & 8.41 ms   &92.10    & 89.84  \\
\B \myNAS-12 & 3.89 ms & 91.53 & 86.93 & 3.84 ms & 91.49 & 87.26 & 6.83 ms   &92.10    & 88.82  \\
\B \myNAS-10 & 4.21 ms & 92.18 & 88.10 & 4.35 ms & 91.73 & 88.24 & 7.40 ms   &92.14    & 88.90  \\
\B \myNAS-8  & 3.02 ms & 90.11 & 84.63 & 3.12 ms & 90.07 & 84.66 & 6.06 ms   &90.52    & 85.41  \\
\B \myNAS-6  & 2.31 ms & 89.79 & 83.73 & 2.41 ms & 90.23 & 84.90 & 3.92 ms   &90.32    & 84.99  \\\hline
\B  Avg.     & 4.36 ms & 91.39 & 86.77 & 4.40 ms & 91.46 & 87.55 & 7.83 ms & 91.72 & 88.04\\\hline
\end{tabular}
    
    % Labeling
    \label{tab:ablation2}
\end{table}
\section{Conclusion}
We presented an automatic neural architecture search that consider the latency factor in the search.
We designed a loss function that constrains the latency to a specific timing for a given hardware.
We show with empirical results that our architectures \myNAS reach the latency it has been designed for on ModelNet10 and generalize on ModelNet40.
Moreover, we showed transfer capabilities of \myNAS for part segmentation, displaying state-of-the-art results on the PartNet benchmark.
We envision \myNAS to be used in time-constrained applications such as autonomous driving, robotics and embedded systems, where latency is of paramount importance for the fulfillment of the vision task.
We believe our work will pave the ground for further constrained neural architecture search.

% \end{document}
\clearpage
\bibliographystyle{splncs04}
\bibliography{egbib}

\begin{thebibliography}{10}
\providecommand{\url}[1]{\texttt{#1}}
\providecommand{\urlprefix}{URL }
\providecommand{\doi}[1]{https://doi.org/#1}

\bibitem{bender2019understanding}
Bender, G.: Understanding and simplifying one-shot architecture search  (2019)

\bibitem{brock2017smash}
Brock, A., Lim, T., Ritchie, J.M., Weston, N.: Smash: one-shot model
  architecture search through hypernetworks. arXiv preprint arXiv:1708.05344
  (2017)

\bibitem{cai2018proxylessnas}
Cai, H., Zhu, L., Han, S.: Proxylessnas: Direct neural architecture search on
  target task and hardware. arXiv preprint arXiv:1812.00332  (2018)

\bibitem{chen2017deeplab}
Chen, L.C., Papandreou, G., Kokkinos, I., Murphy, K., Yuille, A.L.: Deeplab:
  Semantic image segmentation with deep convolutional nets, atrous convolution,
  and fully connected crfs. IEEE transactions on pattern analysis and machine
  intelligence  \textbf{40}(4),  834--848 (2017)

\bibitem{chen2019progressive}
Chen, X., Xie, L., Wu, J., Tian, Q.: Progressive differentiable architecture
  search: Bridging the depth gap between search and evaluation. arXiv preprint
  arXiv:1904.12760  (2019)

\bibitem{chu2019fairnas}
Chu, X., Zhang, B., Xu, R., Li, J.: Fairnas: Rethinking evaluation fairness of
  weight sharing neural architecture search. arXiv preprint arXiv:1907.01845
  (2019)

\bibitem{courbariaux2015binaryconnect}
Courbariaux, M., Bengio, Y., David, J.P.: Binaryconnect: Training deep neural
  networks with binary weights during propagations. In: Advances in neural
  information processing systems. pp. 3123--3131 (2015)

\bibitem{ding2019votenet}
Ding, Z., Han, X., Niethammer, M.: Votenet: A deep learning label fusion method
  for multi-atlas segmentation. In: International Conference on Medical Image
  Computing and Computer-Assisted Intervention. pp. 202--210. Springer (2019)

\bibitem{dong2019search}
Dong, X., Yang, Y.: Searching for a robust neural architecture in four gpu
  hours. In: Proceedings of the IEEE Conference on Computer Vision and Pattern
  Recognition (CVPR). pp. 1761--1770 (2019)

\bibitem{pc_engelmann2018}
Engelmann, F., Kontogianni, T., Hermans, A., Leibe, B.: Exploring spatial
  context for 3d semantic segmentation of point clouds  (feb 2018)

\bibitem{hamilton2017inductive}
Hamilton, W., Ying, Z., Leskovec, J.: Inductive representation learning on
  large graphs. In: Advances in Neural Information Processing Systems. pp.
  1024--1034 (2017)

\bibitem{he2017mask}
He, K., Gkioxari, G., Doll{\'a}r, P., Girshick, R.: Mask r-cnn. In: Proceedings
  of the IEEE international conference on computer vision. pp. 2961--2969
  (2017)

\bibitem{he2016deep}
He, K., Zhang, X., Ren, S., Sun, J.: Deep residual learning for image
  recognition. In: Proceedings of the IEEE conference on computer vision and
  pattern recognition. pp. 770--778 (2016)

\bibitem{hu2020tf}
Hu, Y., Wu, X., He, R.: Tf-nas: Rethinking three search freedoms of
  latency-constrained differentiable neural architecture search. arXiv preprint
  arXiv:2008.05314  (2020)

\bibitem{huang2017densely}
Huang, G., Liu, Z., Van Der~Maaten, L., Weinberger, K.Q.: Densely connected
  convolutional networks. In: Proceedings of the IEEE conference on computer
  vision and pattern recognition. pp. 4700--4708 (2017)

\bibitem{pc_huang2018recurrent}
Huang, Q., Wang, W., Neumann, U.: Recurrent slice networks for 3d segmentation
  of point clouds. In: Proceedings of the IEEE Conference on Computer Vision
  and Pattern Recognition. pp. 2626--2635 (2018)

\bibitem{kipf2016semi}
Kipf, T.N., Welling, M.: Semi-supervised classification with graph
  convolutional networks. arXiv preprint arXiv:1609.02907  (2016)

\bibitem{krizhevsky2012imagenet}
Krizhevsky, A., Sutskever, I., Hinton, G.E.: Imagenet classification with deep
  convolutional neural networks. In: Advances in neural information processing
  systems. pp. 1097--1105 (2012)

\bibitem{le2020going}
Le, E.T., Kokkinos, I., Mitra, N.J.: Going deeper with lean point networks. In:
  Proceedings of the IEEE/CVF Conference on Computer Vision and Pattern
  Recognition. pp. 9503--9512 (2020)

\bibitem{lecun1998gradient}
LeCun, Y., Bottou, L., Bengio, Y., Haffner, P.: Gradient-based learning applied
  to document recognition. Proceedings of the IEEE  \textbf{86}(11),
  2278--2324 (1998)

\bibitem{Li2019DeepGCNsMG}
Li, G., M{\"u}ller, M., Qian, G., Delgadillo, I.C., Abualshour, A., Thabet,
  A.K., Ghanem, B.: Deepgcns: Making gcns go as deep as cnns. ArXiv
  \textbf{abs/1910.06849} (2019)

\bibitem{Li2019DeepGCNs}
Li, G., Müller, M., Thabet, A., Ghanem, B.: Deepgcns: Can gcns go as deep as
  cnns? In: The IEEE International Conference on Computer Vision (ICCV) (2019)

\bibitem{li2019sgas}
Li, G., Qian, G., Delgadillo, I.C., M{\"u}ller, M., Thabet, A., Ghanem, B.:
  Sgas: Sequential greedy architecture search. In: Proceedings of IEEE
  Conference on Computer Vision and Pattern Recognition (CVPR) (2020)

\bibitem{li2018sonet}
Li, J., Chen, B.M., Lee, G.H.: So-net: Self-organizing network for point cloud
  analysis. arXiv preprint arXiv:1803.04249  (2018)

\bibitem{pc_pointcnn}
Li, Y., Bu, R., Sun, M., Wu, W., Di, X., Chen, B.: Pointcnn: Convolution on
  x-transformed points. In: NeurIPS (2018)

\bibitem{liu2018progressive}
Liu, C., Zoph, B., Neumann, M., Shlens, J., Hua, W., Li, L.J., Fei-Fei, L.,
  Yuille, A., Huang, J., Murphy, K.: Progressive neural architecture search.
  In: Proceedings of the European Conference on Computer Vision (ECCV). pp.
  19--34 (2018)

\bibitem{liu2018darts}
Liu, H., Simonyan, K., Yang, Y.: Darts: Differentiable architecture search.
  arXiv preprint arXiv:1806.09055  (2018)

\bibitem{liu2019rscnn}
Liu, Y., Fan, B., Xiang, S., Pan, C.: Relation-shape convolutional neural
  network for point cloud analysis. In: IEEE Conference on Computer Vision and
  Pattern Recognition (CVPR). pp. 8895--8904 (2019)

\bibitem{ds_partnet}
Mo, K., Zhu, S., Chang, A.X., Yi, L., Tripathi, S., Guibas, L.J., Su, H.:
  Partnet: A large-scale benchmark for fine-grained and hierarchical part-level
  3d object understanding. In: Proceedings of the IEEE Conference on Computer
  Vision and Pattern Recognition. pp. 909--918 (2019)

\bibitem{nayman2019xnas}
Nayman, N., Noy, A., Ridnik, T., Friedman, I., Jin, R., Zelnik-Manor, L.: Xnas:
  Neural architecture search with expert advice. arXiv preprint
  arXiv:1906.08031  (2019)

\bibitem{pham2018efficient}
Pham, H., Guan, M.Y., Zoph, B., Le, Q.V., Dean, J.: Efficient neural
  architecture search via parameter sharing. arXiv preprint arXiv:1802.03268
  (2018)

\bibitem{pham2017column}
Pham, T., Tran, T., Phung, D., Venkatesh, S.: Column networks for collective
  classification. In: Thirty-First AAAI Conference on Artificial Intelligence
  (2017)

\bibitem{pc_qi2017pointnet}
Qi, C.R., Su, H., Mo, K., Guibas, L.J.: Pointnet: Deep learning on point sets
  for 3d classification and segmentation. Proc. Computer Vision and Pattern
  Recognition (CVPR), IEEE  \textbf{1}(2), ~4 (2017)

\bibitem{pc_qi2017pointnetpp}
Qi, C.R., Yi, L., Su, H., Guibas, L.J.: Pointnet++: Deep hierarchical feature
  learning on point sets in a metric space. In: Advances in Neural Information
  Processing Systems. pp. 5099--5108 (2017)

\bibitem{ren2015faster}
Ren, S., He, K., Girshick, R., Sun, J.: Faster r-cnn: Towards real-time object
  detection with region proposal networks. In: Advances in neural information
  processing systems. pp. 91--99 (2015)

\bibitem{ronneberger2015u}
Ronneberger, O., Fischer, P., Brox, T.: U-net: Convolutional networks for
  biomedical image segmentation. In: International Conference on Medical image
  computing and computer-assisted intervention. pp. 234--241. Springer (2015)

\bibitem{chi2019prcnn}
Shi, S., Wang, X., Li, H.: Pointrcnn: 3d object proposal generation and
  detection from point cloud. In: The IEEE Conference on Computer Vision and
  Pattern Recognition (CVPR) (June 2019)

\bibitem{simonyan2014very}
Simonyan, K., Zisserman, A.: Very deep convolutional networks for large-scale
  image recognition. arXiv preprint arXiv:1409.1556  (2014)

\bibitem{stamoulis2019single}
Stamoulis, D., Ding, R., Wang, D., Lymberopoulos, D., Priyantha, B., Liu, J.,
  Marculescu, D.: Single-path nas: Designing hardware-efficient convnets in
  less than 4 hours. arXiv preprint arXiv:1904.02877  (2019)

\bibitem{GoogLeNet2015}
Szegedy, C., Liu, W., Jia, Y., Sermanet, P., Reed, S., Anguelov, D., Erhan, D.,
  Vanhoucke, V., Rabinovich, A.: Going deeper with convolutions. In:
  Proceedings of the IEEE conference on computer vision and pattern
  recognition. pp.~1--9 (2015)

\bibitem{szegedy2016rethinking}
Szegedy, C., Vanhoucke, V., Ioffe, S., Shlens, J., Wojna, Z.: Rethinking the
  inception architecture for computer vision. In: Proceedings of the IEEE
  conference on computer vision and pattern recognition. pp. 2818--2826 (2016)

\bibitem{tan2019mnasnet}
Tan, M., Chen, B., Pang, R., Vasudevan, V., Sandler, M., Howard, A., Le, Q.V.:
  Mnasnet: Platform-aware neural architecture search for mobile. In:
  Proceedings of the IEEE Conference on Computer Vision and Pattern
  Recognition. pp. 2820--2828 (2019)

\bibitem{tang2020searching}
Tang, H., Liu, Z., Zhao, S., Lin, Y., Lin, J., Wang, H., Han, S.: Searching
  efficient 3d architectures with sparse point-voxel convolution. arXiv
  preprint arXiv:2007.16100  (2020)

\bibitem{pc_tatarchenko2018tangent}
Tatarchenko, M., Park, J., Koltun, V., Zhou, Q.Y.: Tangent convolutions for
  dense prediction in 3d. In: Proceedings of the IEEE Conference on Computer
  Vision and Pattern Recognition. pp. 3887--3896 (2018)

\bibitem{pc_kpconv}
Thomas, H., Qi, C.R., Deschaud, J.E., Marcotegui, B., Goulette, F., Guibas,
  L.J.: Kpconv: Flexible and deformable convolution for point clouds. ArXiv
  \textbf{abs/1904.08889} (2019)

\bibitem{velivckovic2017graph}
Veli{\v{c}}kovi{\'c}, P., Cucurull, G., Casanova, A., Romero, A., Lio, P.,
  Bengio, Y.: Graph attention networks. arXiv preprint arXiv:1710.10903  (2017)

\bibitem{dgcnn}
Wang, Y., Sun, Y., Liu, Z., Sarma, S.E., Bronstein, M.M., Solomon, J.M.:
  Dynamic graph cnn for learning on point clouds. ACM Transactions on Graphics
  (TOG)  (2019)

\bibitem{wu2019fbnet}
Wu, B., Dai, X., Zhang, P., Wang, Y., Sun, F., Wu, Y., Tian, Y., Vajda, P.,
  Jia, Y., Keutzer, K.: Fbnet: Hardware-aware efficient convnet design via
  differentiable neural architecture search. In: Proceedings of the IEEE
  Conference on Computer Vision and Pattern Recognition. pp. 10734--10742
  (2019)

\bibitem{ds_modelnet}
Wu, Z., Song, S., Khosla, A., Yu, F., Zhang, L., Tang, X., Xiao, J.: 3d
  shapenets: A deep representation for volumetric shapes. 2015 IEEE Conference
  on Computer Vision and Pattern Recognition (CVPR) pp. 1912--1920 (2014)

\bibitem{xie2018snas}
Xie, S., Zheng, H., Liu, C., Lin, L.: Snas: stochastic neural architecture
  search. arXiv preprint arXiv:1812.09926  (2018)

\bibitem{Xu2018HowPAGIN}
Xu, K., Hu, W., Leskovec, J., Jegelka, S.: How powerful are graph neural
  networks? ArXiv  \textbf{abs/1810.00826} (2018)

\bibitem{xu2020gtad}
Xu, M., Zhao, C., Rojas, D.S., Thabet, A., Ghanem, B.: G-tad: Sub-graph
  localization for temporal action detection (June 2020)

\bibitem{pc_xu2018spidercnn}
Xu, Y., Fan, T., Xu, M., Zeng, L., Qiao, Y.: Spidercnn: Deep learning on point
  sets with parameterized convolutional filters. In: Proceedings of the
  European Conference on Computer Vision (ECCV). pp. 87--102 (2018)

\bibitem{xu2019pc}
Xu, Y., Xie, L., Zhang, X., Chen, X., Qi, G.J., Tian, Q., Xiong, H.: Pc-darts:
  Partial channel connections for memory-efficient differentiable architecture
  search. arXiv preprint arXiv:1907.05737  (2019)

\bibitem{xu2020latency}
Xu, Y., Xie, L., Zhang, X., Chen, X., Shi, B., Tian, Q., Xiong, H.:
  Latency-aware differentiable neural architecture search. arXiv preprint
  arXiv:2001.06392  (2020)

\bibitem{pc_ye20183d}
Ye, X., Li, J., Huang, H., Du, L., Zhang, X.: 3d recurrent neural networks with
  context fusion for point cloud semantic segmentation. In: European Conference
  on Computer Vision. pp. 415--430. Springer (2018)

\bibitem{pc_shellnet}
Zhang, Z., Hua, B.S., Yeung, S.K.: Shellnet: Efficient point cloud
  convolutional neural networks using concentric shells statistics. ArXiv
  \textbf{abs/1908.06295} (2019)

\bibitem{zhou2019bayesnas}
Zhou, H., Yang, M., Wang, J., Pan, W.: Bayesnas: A bayesian approach for neural
  architecture search (2019)

\bibitem{zoph2016neural}
Zoph, B., Le, Q.V.: Neural architecture search with reinforcement learning.
  arXiv preprint arXiv:1611.01578  (2016)

\bibitem{zoph2018learning}
Zoph, B., Vasudevan, V., Shlens, J., Le, Q.V.: Learning transferable
  architectures for scalable image recognition. In: Proceedings of the IEEE
  conference on computer vision and pattern recognition. pp. 8697--8710 (2018)

\end{thebibliography}

\clearpage
\appendix
\section{Visualization}
To optimize the model architecture $\mathcal{A}$ with the latency constraint in an end-to-end fashion, we need the gradient with respect to the binary architecture encoding  $\mathbf{\tilde{E}}=\mathcal{E}(\mathcal{A})$ . However, we compute $\mathbf{\tilde{E}}$ from a non-differentiable function $\mathcal{E}(\cdot)$ that involves some rules/heuristics that cannot directly retrieve the gradients. Hence, we approximate the gradient by trusting the selection rules/heuristics as a linear operation from Eq.(3). 

\begin{figure}[h]
    \centering
    \includegraphics[width=1\textwidth,trim={0cm 0.3cm 0cm 0.5cm},clip]{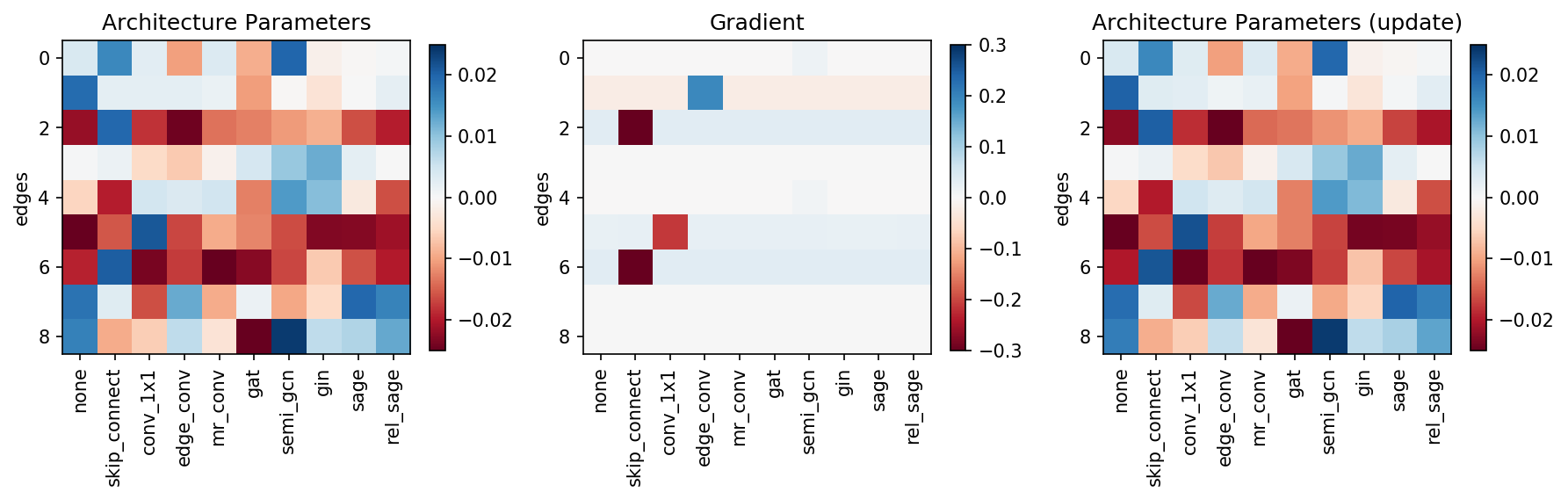}
    \caption{\textbf{Visualization of  the updating procedure in one training iteration.} The left and right plots show the architecture parameters before and after one gradient step respectively. The middle plot shows the gradient from the back-propagation on the latency loss. 
    }
    \label{fig:vis_iter}
\end{figure}

% \noindent
% \subsection{Visualization of One Training Iteration}
We visualize the updating procedure of one training iteration in Fig.~\ref{fig:vis_iter}. The first plot shows the input architecture parameters before  this iteration. Besides, the middle plot shows its gradient from the back-propagation on the latency loss. In the last plot, we compare the updated architecture parameters. As explained in paper Section 3.2,  the element $(m,n)$ in the plots indicates that the operation $n$ is chosen for the edge $m$. Notably, the second row, \textit{EdgeConv} column in the Gradient plot is positive, indicating that decreasing the weight of the \textit{EdgeConv} operation on the second edge will decrease the latency. In the right plot, accordingly, we decrease the weight of the \textit{EdgeConv} operator on the this edge.

\begin{figure}[h]
    \centering
    \includegraphics[width=1\textwidth,trim={0cm 0.3cm 0cm 0.5cm},clip]{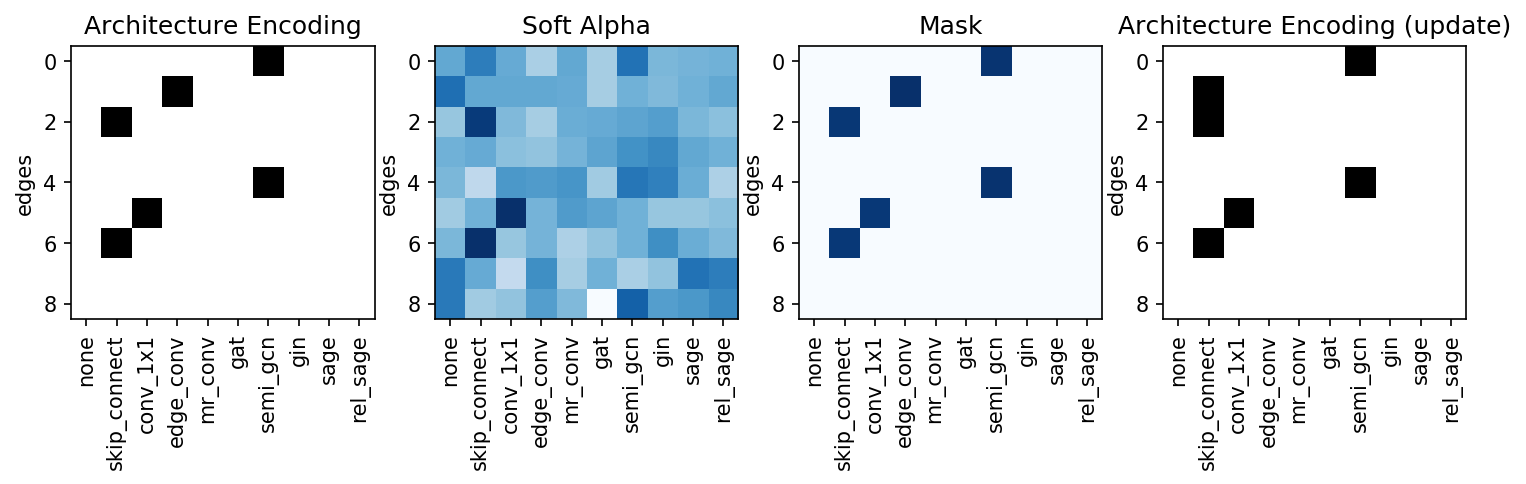}
    \caption{\textbf{Visualization of the intermediate variables in training.} The four plots repectively show the input architecture encoding, softmax result, mask, and the updated architecture encoding. Eventually, the \textit{EdgeConv} operation on the second edge is replaced by a more efficient \textit{Skip-Connect} operation.
    }
    \label{fig:vis_var}
\end{figure}

\begin{figure}[h]
\centering
\begin{tabular}{@{}c@{\hspace{1mm}}c@{\hspace{1mm}}c@{}}
\includegraphics[width=0.5\columnwidth,trim={0cm 1cm 0cm 1cm},clip]{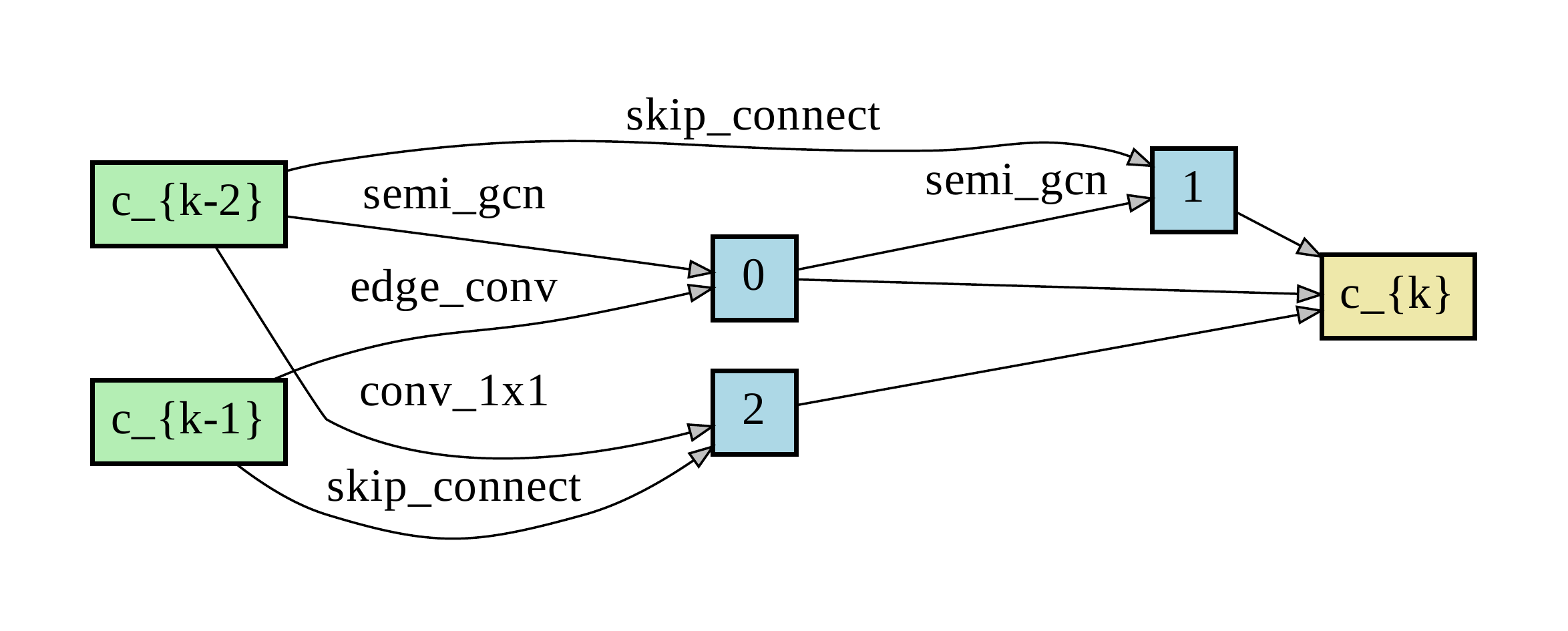} 
\includegraphics[width=0.5\columnwidth,trim={0cm 1cm 0cm 1cm},clip]{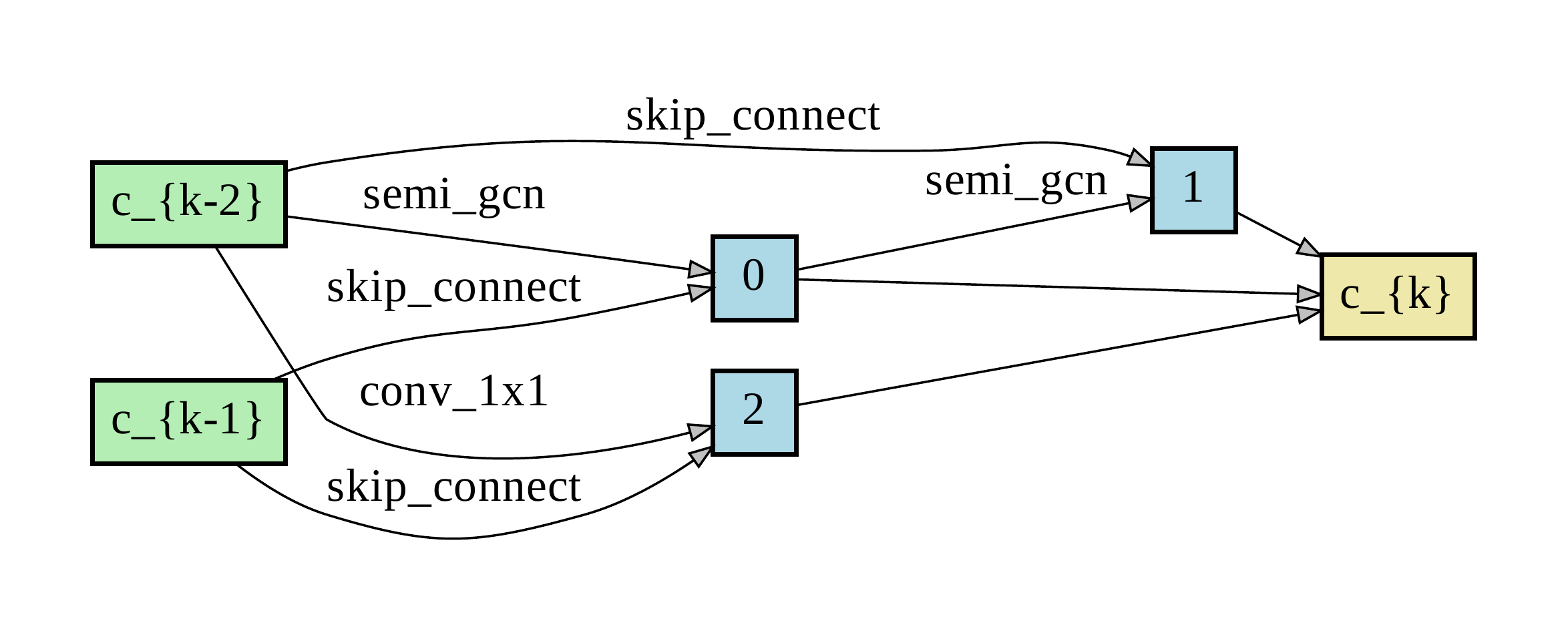}
\end{tabular}
\caption{\textbf{Visualization of discovered architectures before (top) and after (bottom) one iteration.} Please be note that the edge from $c_{k-1}$ to node $0$ changes from \textit{EdgeConv} to more efficient \textit{Skip-Connect}.}
\label{fig:vis_iter_compare}
\end{figure}

Fig.~\ref{fig:vis_var} visualizes the intermediate variables. Given the architecture parameter $\mathcal{A}$, we use some rules/heuristics to compute its encoding $\mathbf{\tilde{E}}$, shown in the first plot. Meantime, we apply \textit{softmax} on each row of $\mathcal{A}$ to get the second plot. Then, to approximate the gradient, we trust the selection rules/heuristics as linear operation by approximating with multiplying an element-wise mask $\bm{\zeta}$, shown in the third plot. That is to say, the element-wise multiplication of Soft Alpha and Mask equals to the Architecture Encoding. Finally, the architecture parameters are updated and result in an new encoding in the last plot, where the \textit{EdgeConv} operation is replaced by a more efficient \textit{Skip-Connect} operation. We also compare the changes of cells in Fig.~\ref{fig:vis_iter_compare}.

% \newpage
\subsection{More Gradient Visualization}
\begin{figure}[h]
    \centering
    \includegraphics[width=0.8\textwidth,trim={0cm 0.3cm 0cm 1cm},clip]{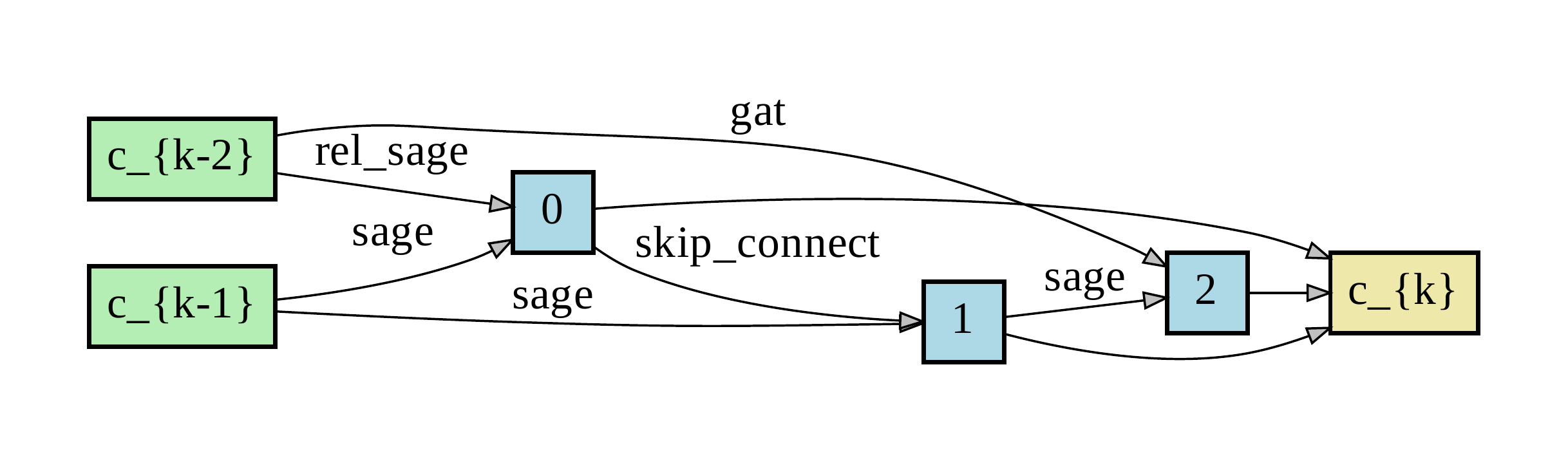}
    \caption{\textbf{Visualization of a random selected architecture in the test set.} Its measured latency and predicted latency are 13.92ms and 14.33ms respectively.}
    \label{fig:arch_test}
\end{figure}

To give a comprehensive study on the quality of the gradient approximation, we randomly choose one model architecture in the test set (shown in Fig.~\ref{fig:arch_test}), and visualize a approximation with different setups. The measured latency and predicted latency of this architecture are respectively 13.92ms and 14.33ms.

\begin{figure}[!h]
    \centering
    \includegraphics[width=0.8\textwidth,]{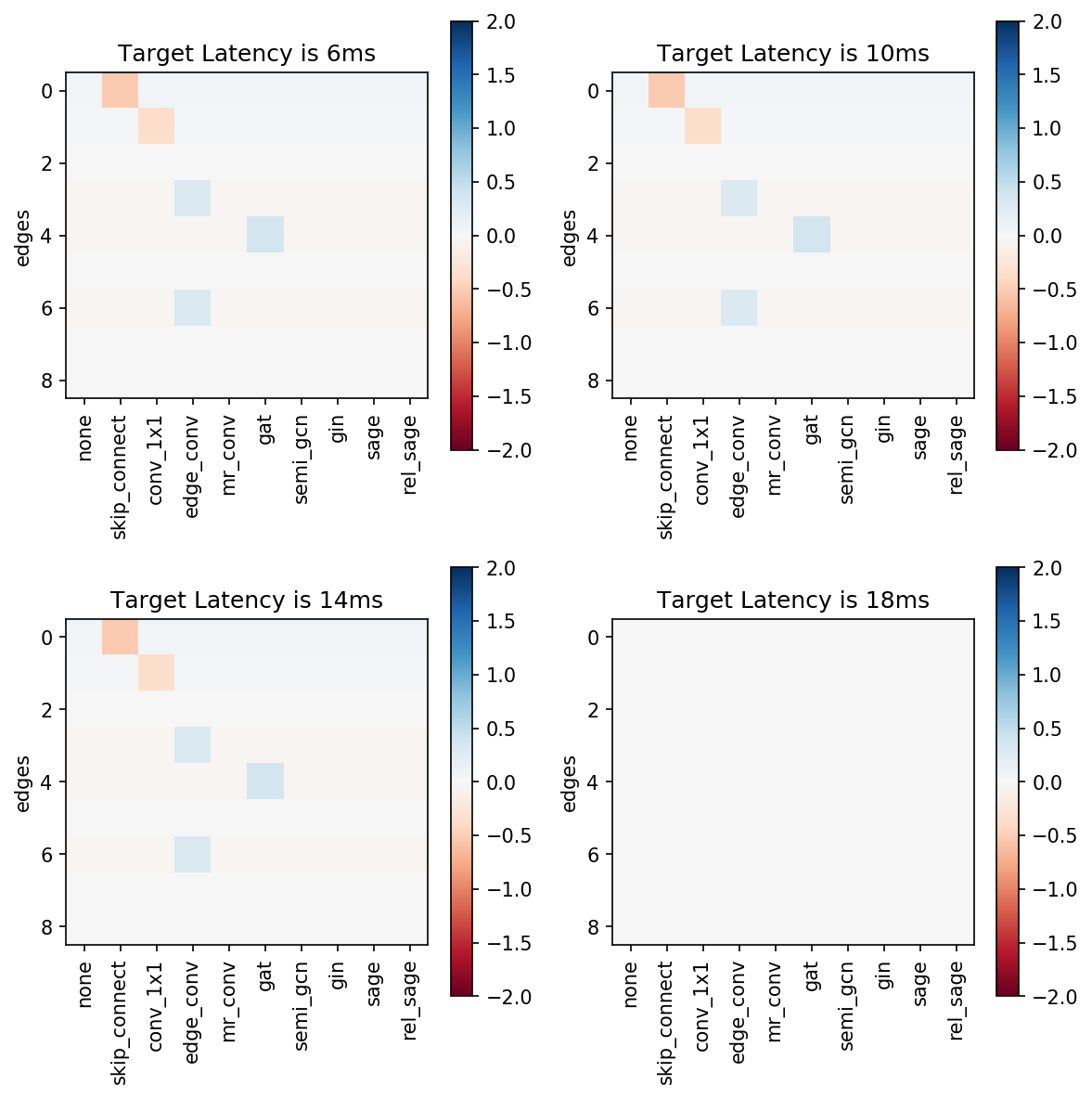}
    \caption{\textbf{Visualization of the gradient from latency constraint (1).} We apply hinge-loss-like regularization loss to regress the \textbf{LatReg} model prediction to four target latencies. }
    \label{fig:vis_grad1}
\end{figure}

% \newpage
\mysection{Gradient Visualization with Hinge Latency loss}
We visualize the gradient in Fig.~\ref{fig:vis_grad1}, where we use a hinge loss  in the latency constraint.  When we set the latency targets as 6ms, 10ms, or 14ms, the gradients stay the same. Moreover, the negative gradient direction tends to assign higher value for efficient operations such as \textit{Skip-Connect} and \textit{Conv-1$\times$1}, while decrease the value for complex GCN operations like \textit{MRConv} and \textit{EdgeConv}. Besides, when the target latency is 18ms, which is greater to the current model latency 14.33ms, the gradients of the hinge latency loss become zeros.

\begin{figure}[h]
    \centering
    \includegraphics[width=0.8\textwidth,]{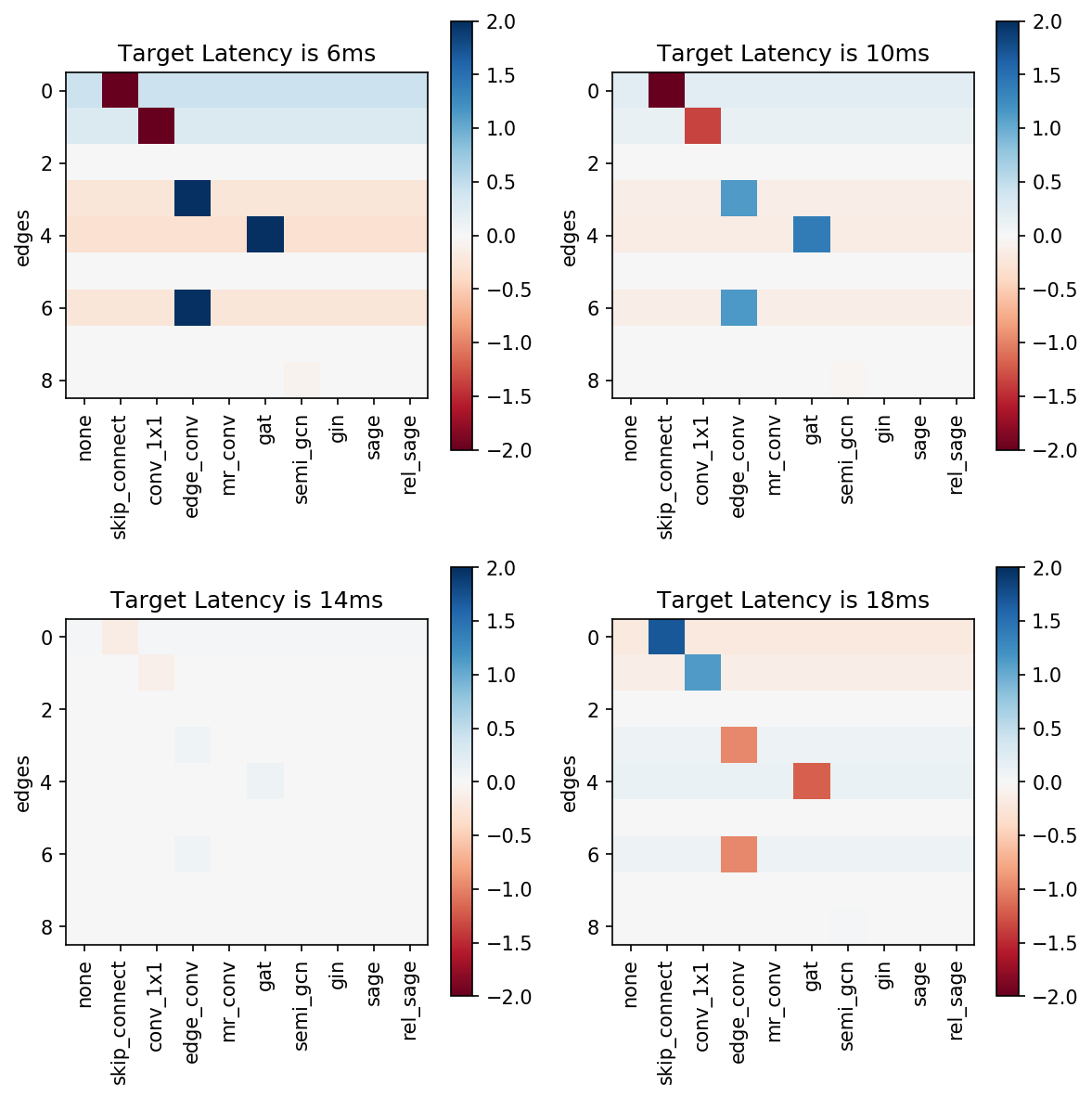}
    \caption{\textbf{Visualization of the gradient from latency constraint (2).} We apply MSE loss to regress the \textbf{LatReg} model prediction to four target latencies.}
    \label{fig:vis_grad2}
\end{figure}

\mysection{Gradient Visualization with MSE Latency loss}
We also visualize the gradient in
Fig.~\ref{fig:vis_grad2}, where we use a mean-square-error (MSE) loss in the latency constraint.  When we set the latency targets as 6ms, 10ms, and 14ms, the gradients have the same pattern, but the scales are positively correlated the latency gap between targets and the prediction (14.33ms). Notably, when the target latency is greater to the current model latency, shown in the bottom-right plot, the negative gradient will assign less probability for \textit{Skip-Connect} and \textit{Conv-1$\times$1} (the cheap operations), while increase the probability for time-consuming operations such as \textit{MRConv} and \textit{EdgeConv}.

\newpage
\section{Details in Latency Regressor Model}
\begin{figure}[ht]
    \centering
    \includegraphics[width=0.5\textwidth]{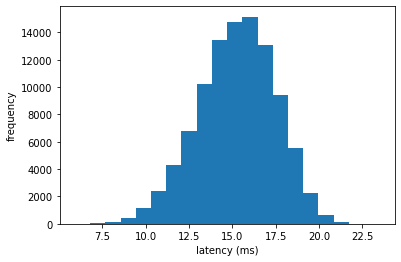}
    \includegraphics[width=0.47\textwidth]{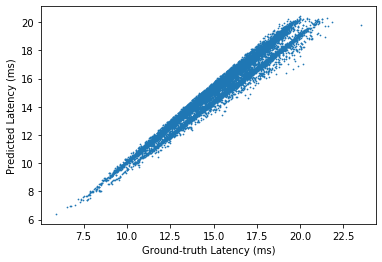}
    \caption{\textbf{LatReg Model data distribution and performance.} \textit{Left}: We show here the distribution of all latencies. Latency values range from $5.9ms$ to $23.5ms$, with an average around $15ms$. We randomly split out architectures into train, validation, and test subsets, thus keeping similar distributions on all splits.\textit{Right}: We can find the the measured latency and predicted latancy are linearly dependent to each other and the slope is close to 1.}
    \label{fig:lat_dist}
\end{figure}

To train our Latency Regressor (LatReg) model, we sample $100K$ random cell-architectures from our search space and measure their latencies. We show the latency distribution in \figLabel \ref{fig:lat_dist} \textit{Left}.

We randomly split our architectures into train, validation, and test sets. We use the training set to train our LatReg model, use the validation set to choose the hyper-parameters such as learning rate and training epochs. Then, we evaluate the model on the test set, and the LatReg model reaches an average absolute error of 0.16ms. \figLabel \ref{fig:lat_dist} \textit{Right} visualized the measured latency and predicted latancy of each architecture in the test set. We can find the two variables are linearly dependent to each other and the slope is close to 1.

\section{Ablation Study on Non-targeted Latency Loss}
\begin{figure}[!h]
 \centering
 \begin{tabular}{@{}c@{\hspace{1mm}}c@{\hspace{1mm}}c@{}}
 		
 		\includegraphics[width=0.5\columnwidth]{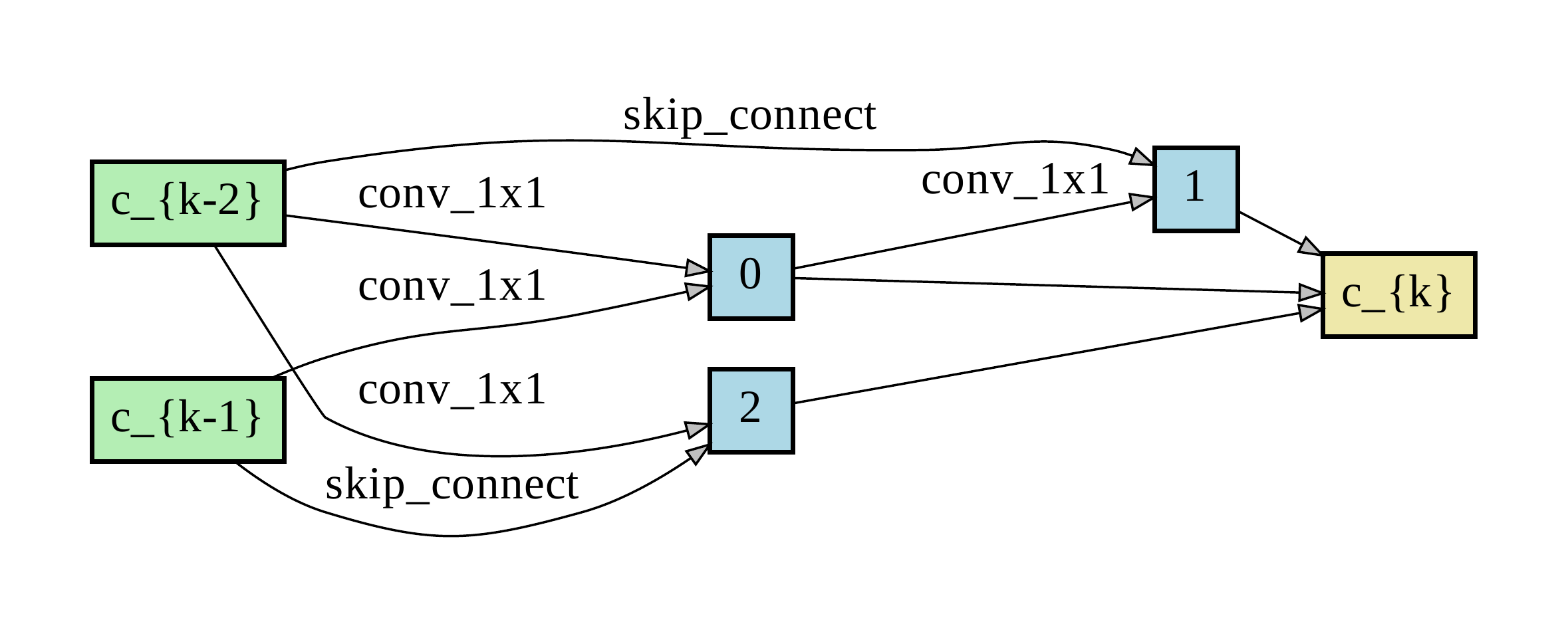}
 		 &
 		\includegraphics[width=0.5\columnwidth]{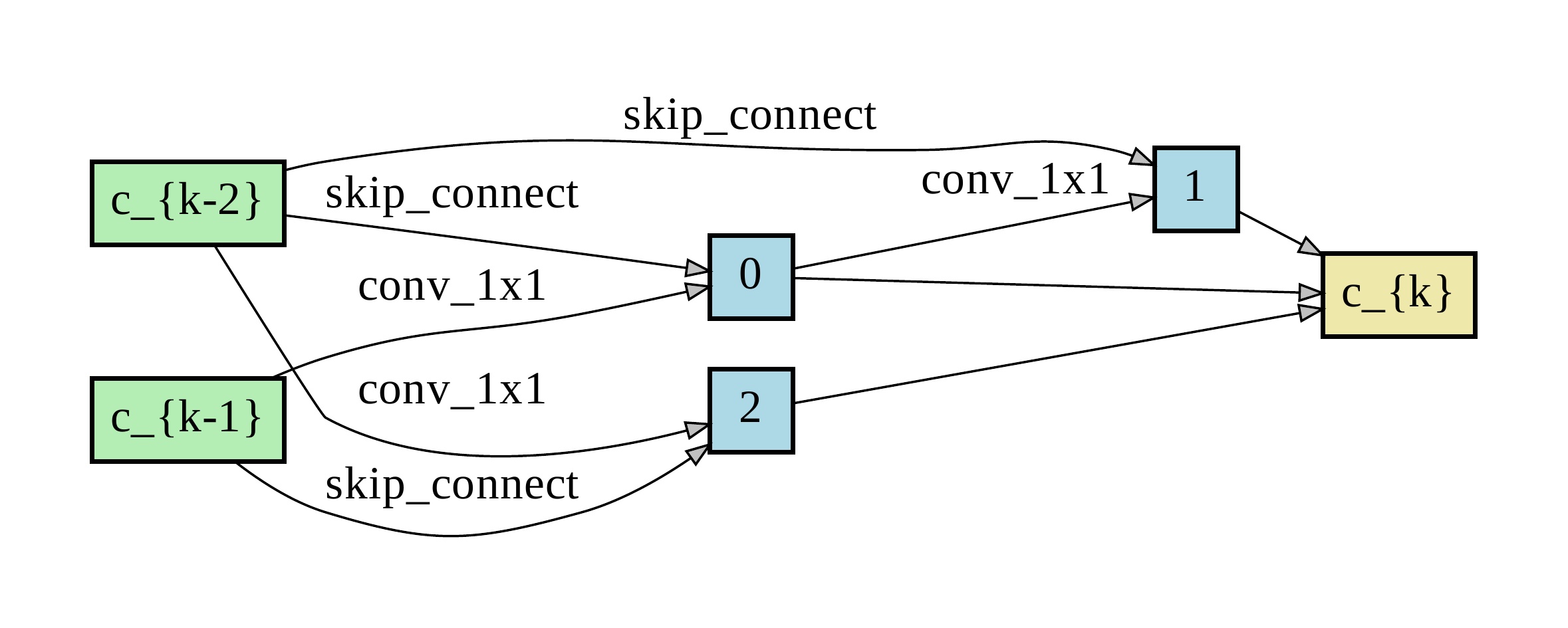}
		 \\
 		
 		\small (a) $\lambda = 0.5$ & \small (b) $\lambda = 0.1$ \\
 		
 		\includegraphics[width=0.5\columnwidth]{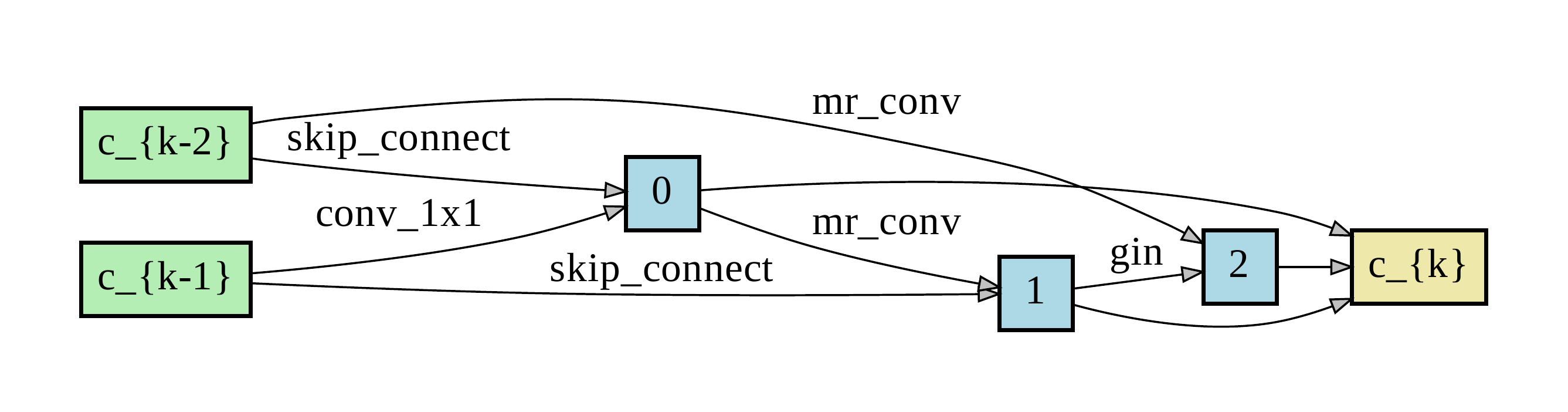} 
 		 &
        \includegraphics[width=0.5\columnwidth]{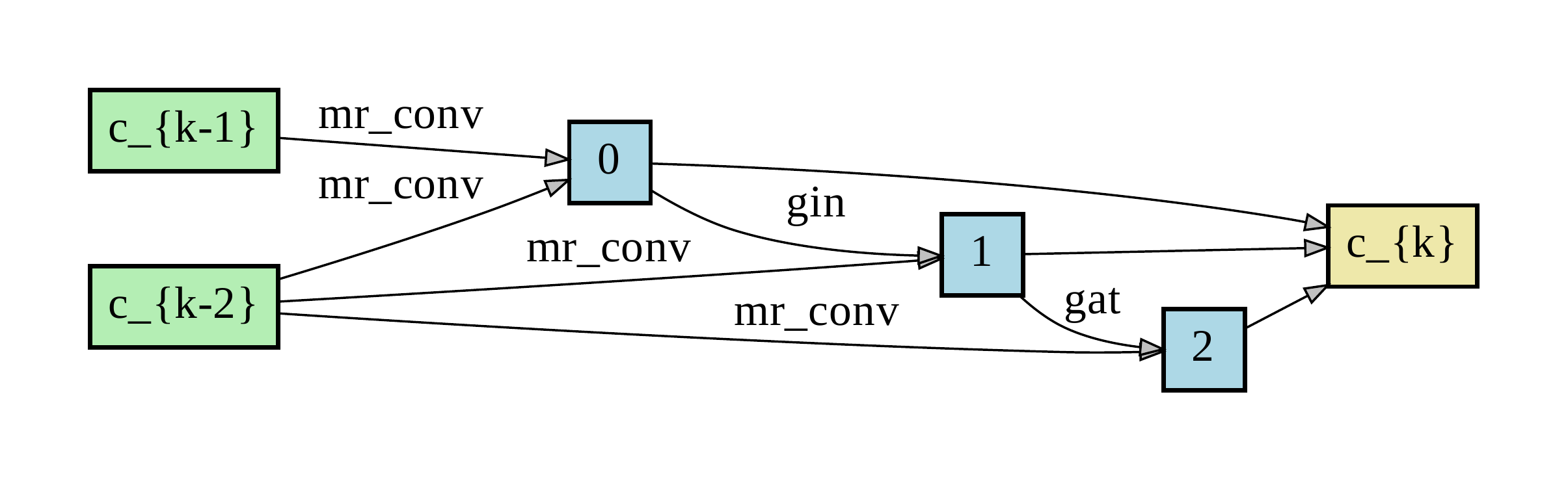}
		 \\
		
		\small (c) $\lambda = 0.0005$ & \small (d) $\lambda = 0.0001$ \\

 \end{tabular}
 \caption{Visualization of searched cells with non-targeted latency loss with different $\lambda$.}
 \label{fig:visualizations_gcn_nt}
 \end{figure}

Previous latency-aware differentiable NAS methods \cite{cai2018proxylessnas,wu2019fbnet} usually add the latency loss as a regularizer to the bi-level optimization procedure as follows:
\begin{align}
	\min_\mathcal{A} \quad & \mathcal{L}_{val}(\mathcal{W}^*(\mathcal{A}), \mathcal{A}) {~\color{blue} + \lambda LatReg(\mathcal{E}(\mathcal{A})) } \label{eq:outer} \\
	\text{s.t.} \quad &\mathcal{W}^*(\mathcal{A}) = \mathrm{argmin}_\mathcal{W} \enskip \mathcal{L}_{train}(\mathcal{W}, \mathcal{A}) \label{eq:inner}
\end{align}
where $\mathcal{L}_{val}$ is the cross-entropy loss on validation set, $\mathcal{L}_{train}$ is the cross-entropy loss on training set, $\mathcal{A}$ is the architectural parameters, $\mathcal{W}$ is the network weights, $\lambda$ is the regularization factor, $LatReg(\cdot)$ is the learned latency regressor and $\mathcal{E}(\cdot)$ is a non-differentiable binarized function that take as input continuous architectural parameters $\mathcal{A}$ and output a binarized architecture encoding $\mathbf{\tilde{E}}$. This non-targeted latency loss can be regarded as a special case of the proposed hinge latency loss with a target as $0$. As we discussed, this form of constrained loss has two main disadvantages 
(1) It fails to minimize the architecture to be lower than a certain latency. Thus, it is less controllable while using for hardware deployment.
(2) The regularization factor $\lambda$ is sensitive and difficult to tune to trade-off the capacity and efficiency of searched models. In \tblLabel \ref{tab:non_target}, we experiment non-targeted latency loss with $\lambda$ from $0.5$ to $0.0001$. We first set $\lambda = 0.5$. The obtained architecture only consists of \textit{Skip-Connect}, \textit{Conv-1$\times$1} (see \figLabel \ref{fig:visualizations_gcn_nt} (a)) and merely reach $90.24\%$ overall accuracy on ModelNet40. We further decrease the regularization factor $\lambda$ to $0.1$ and $0.05$. The discovered cells still only consist of \textit{Skip-Connect}, \textit{Conv-1$\times$1}. Therefore, the regularization are still considered too strong. Only when $\lambda$ is not greater than $0.001$, the searched cells have reasonable performance around $92.5\%$. However, the value of $\lambda$ does not explicitly guarantee a certain latency constraint which make it hard to tune. The searched cells are visualized in \figLabel \ref{fig:visualizations_gcn_nt}.

\begin{table}[h]
    \centering
    \small
    \caption{\textbf{Ablation on Non-Targeted Latency Loss.} We change the regularization factor $\lambda$ from $0.5$ to $0.0001$. We see that the performance of the discovered networks on ModelNet40 are sensitive to $\lambda$ and the latency of networks is hard to control.}
\begin{tabular}{l||c|c|c}
\B $\bm{\lambda}$ & \B Latency & \B O.A. & \B C.A.\\\hline
 
\B 0.5 & 6.60 ms & 90.24 & 84.70 \\
\B 0.1 & 6.19 ms & 90.76 & 85.37 \\
\B 0.05 & 6.35 ms & 90.28 & 84.90 \\
\B 0.01 & 9.64 ms & 92.26 & 89.00 \\
\B 0.005 & 8.82 ms & 86.83 & 80.65 \\
\B 0.001  & 14.63 ms & 92.63 & 89.98 \\
\B 0.0005  & 12.08 ms & 92.50 & 89.75 \\
\B 0.0001  & 18.71 ms & 92.50 & 89.68 \\\hline
\end{tabular}
    
    % Labeling
    \label{tab:non_target}
\end{table}

 \vspace{-15pt}
\end{document}